\documentclass{article}

\usepackage{arxiv}
\usepackage{amsmath}
\usepackage{amssymb}
\usepackage{bbm}
\usepackage{epsfig}
\usepackage{graphics,amsopn,amstext,color,enumerate,bm}
\usepackage{color,float,epstopdf}
\usepackage{mathtools}
\usepackage{svg}
\usepackage{multirow}
\usepackage{graphicx}
\usepackage{natbib}
\usepackage{url}
\usepackage{microtype}
\usepackage{subfigure}
\usepackage{booktabs} 
\usepackage{hyperref}
\usepackage{wrapfig}
\usepackage{adjustbox}
\usepackage[normalem]{ulem}
\useunder{\uline}{\ul}{}

\newcommand{\ba}{{\mathbf{a}}}
\newcommand{\bb}{{\mathbf{b}}}

\newcommand{\bs}{{\mathbf{s}}}

\newcommand{\bv}{{\mathbf{v}}}

\newcommand{\bx}{{\mathbf{x}}}
\newcommand{\by}{{\mathbf{y}}}
\newcommand{\bz}{{\mathbf{z}}}

\newcommand{\btheta}{{\bm{\theta}}}

\newcommand{\Renyi}{R\'enyi }
\newcommand{\Tr}{\textrm{Tr}}
\newcommand{\expec}{\mathbb{E}}

\usepackage{cleveref}
\crefname{algorithm}{Algorithm}{Algorithms}
\crefname{assumption}{Assumption}{Assumptions}
\crefname{equation}{Eq.}{Eqs.}
\crefname{figure}{Fig.}{Figs.}
\crefname{table}{Table}{Tables}
\crefname{section}{Sec.}{Secs.}
\crefname{theorem}{Theorem}{Theorems}
\crefname{lemma}{Lemma}{Lemmas}
\crefname{proposition}{Proposition}{Propositions}
\crefname{definition}{Definition}{Definitions}
\crefname{corollary}{Corollary}{Corollaries}
\crefname{remark}{Remark}{Remarks}
\crefname{example}{Example}{Examples}
\crefname{appendix}{Appendix}{Appendices}

\DeclareMathOperator*{\argmax}{argmax}

\usepackage{algpseudocode, algorithm}
\usepackage[noend]{algcompatible}

\newcommand{\BlackBox}{\rule{1.5ex}{1.5ex}}  

\newenvironment{proof}{\par\noindent{\bf Proof\ }}{\hfill\BlackBox\\[2mm]}
 
\newtheorem{theorem}{Theorem}
\newtheorem{lemma}[theorem]{Lemma} 
\newtheorem{proposition}[theorem]{Proposition} 
\newtheorem{remark}[theorem]{Remark}

\title{Dr. FERMI: A Stochastic Distributionally Robust Fair Empirical Risk Minimization Framework}

\author{
  Sina Baharlouei
  \thanks{Industrial and Systems Engineering Department, University of Southern, California, Los Angeles, CA 90089, USA}
  \\\texttt{baharlou@usc.edu} \\
   \And
  Meisam Razaviyayn
  \footnotemark[1]\\
  \texttt{razaviya@usc.edu} \\
  }

\begin{document}
\maketitle

\begin{abstract}
While training fair machine learning models has been studied extensively in recent years, most developed methods rely on the assumption that the training and test data have similar distributions. 
In the presence of distribution shifts, fair models may behave unfairly on test data. 
There have been some developments for fair learning robust to distribution shifts to address this shortcoming. However, most proposed solutions are based on the assumption of having access to the causal graph describing the interaction of different features. Moreover, existing algorithms require full access to data and cannot be used when small batches are used (stochastic/batch implementation).
This paper proposes the first stochastic distributionally robust fairness framework with convergence guarantees that do not require knowledge of the causal graph.  
More specifically, we formulate the fair inference in the presence of the distribution shift as a distributionally robust optimization problem under $L_p$ norm uncertainty sets with respect to the Exponential \Renyi Mutual Information (ERMI) as the measure of fairness violation. 
We then discuss how the proposed method can be implemented in a stochastic fashion. 
We have evaluated the presented framework's performance and efficiency through extensive experiments on real datasets consisting of distribution shifts. 
\end{abstract}

\vspace{-5mm}
\section{Introduction}
\vspace{-3mm}
\label{sec:introduction}
Machine Learning models demonstrate remarkable results in automated decision-making tasks such as image processing~\citep{krizhevsky2017imagenet}, object detection~\citep{tan2020efficientdet}, natural language understanding~\citep{devlin2018bert, saravani2021automated}, speech recognition~\citep{abdel2014convolutional}, and automated code generation~\citep{vaithilingam2022expectation, nazari}. However, na\"ively optimizing the accuracy/performance may lead to biased models against protected groups such as racial minorities and women~\citep{angwin2016machine, buolamwini2018gender}. 
A wide range of algorithms has been proposed to enhance the \textit{fairness} of machine learning models. These algorithms can be divided into three main categories: pre-processing, post-processing, and in-processing. Pre-processing algorithms remove the dependence/correlation of the training data with the sensitive attributes by transforming the data to a new ``fair representation" in a pre-processing stage~\citep{zemel2013learning,creager2019flexibly}. Post-processing algorithms adjust the decision boundary of the learned model after the training stage to satisfy the fairness constraints of the problem~\citep{hardt2016equality, lohia2019bias, kim2019multiaccuracy}. Finally, in-processing approaches optimize the parameters of the model to maximize accuracy while satisfying the fairness constraints during the training procedure; see, e.g.,~\citep{zafar2017fairness,donini2018empirical, mary2019fairness, baharlouei2019renyi}. 
\vspace{-1mm}

The underlying assumption of most fair learning algorithms is that the train and test domains have an identical distribution. Thus, establishing fairness in the training phase can guarantee fairness in the test phase. However, this assumption does not hold in many real applications. As an example, consider the new collection of datasets (ACS PUMS) released by~\citep{ding2021retiring} with the underlying task of predicting the income level of individuals (similar to the adult dataset~\citep{Dua:2019}). The data points are divided based on different years and states. When several fair algorithms are applied to ACS PUMS dataset to learn fair logistic regression and XGBoost classifiers in the people of one US state, the performance of the learned models is severely dropped in terms of both accuracy and fairness violation when they are evaluated in other states~\citep{ding2021retiring}. We also made a similar observation in our numerical experiments (see Figure~\ref{fig: acs}). As another example,~\citet{schrouff2022maintaining} reviews several examples from the field of healthcare and medicine where a trained ``fair" model in one hospital does not behave fairly in other hospitals. Such examples demonstrate the importance of developing efficient algorithms for training fair models against distribution shifts.
\vspace{-1mm}

Another crucial requirement in modern machine learning algorithms is the amenability to stochastic optimization. In large-scale learning tasks, only a mini-batch of data is used at each algorithm step. Thus, implementing the algorithm based on the mini-batches of data has to improve the loss function over iterates and be convergent.  While this requirement is met when training based on vanilla ERM loss functions~\citep{lee2019first, kale2022gradient}, the stochastic algorithms (such as stochastic gradient descent) may not be necessarily convergent in the presence of fairness regularizers~\citep{lowy2021fermi}.

\vspace{-1mm}

In this work, we propose a distributionally robust optimization framework to maintain fairness across different domains in the presence of distribution shifts. Our framework \textbf{does not rely on having access to the knowledge of the causal graph of features}. Moreover, our framework is \textbf{amenable to stochastic optimization} and comes with convergence guarantees. 

\vspace{-5mm}
\subsection{Related Work}\label{subsec:Related_works}
\vspace{-3mm}
Machine learning models can suffer from severe performance drop when the distribution of the test data differs from training data~\citep{ben2006analysis, moreno2012unifying, recht2018cifar, recht2019imagenet}. 
Learning based on spurious features~\citep{buolamwini2018gender, zhou2021examining}, class imbalance~\citep{li2019overfitting, jing2021towards}, non-ignorable missing values~\citep{xu2009robustness}, and overfitting~\citep{tzeng2014deep} are several factors contributing to such a poor performance on test data. To mitigate distribution shift, \citet{ben2010theory} formalizes the problem as a domain adaptation task where the unlabeled test data is available during the training procedure. In this case, the key idea is to regularize the empirical risk minimization over the given samples via a divergence function measuring the distribution distance of the source and target predictors. 
When data from the target domain is not available, the most common strategy for handling distribution shift is distributionally robust optimization~\citep{rahimian2019distributionally, kuhn2019wasserstein, lin2022distributionally}. This approach relies on minimizing the risk for the worst-case distribution within an uncertainty set around the training data distribution. Such uncertainty sets can be defined as $\ell_1$~\citep{ben2010theory}, $\ell_2$~\citep{gao2017distributionally}, $\ell_{\infty}$~\citep{rifle}, optimal transport~\citep{blanchet2019data}, Wasserstein distance~\citep{kuhn2019wasserstein}, Sinkhorn divergences~\citep{oneto2020exploiting}, $\chi^2$-divergence~\citep{levy2020large}, or Conditional Value at Risk (CVaR)~\citep{zymler2013distributionally, levy2020large} balls around the empirical distribution of the source domain. 
\vspace{-1mm}

A diverse set of methods and analyses are introduced in the context of learning fair models in the presence of the distribution shift. \citet{lechner2021impossibility} shows how learning a \textit{fair representation} using a fair pre-processing algorithm is almost \textit{impossible} even under the demographic parity notion when the distribution of the test data shifts. \citet{dai2020label} empirically demonstrates how bias in labels and changes in the distribution of labels in the target domain can lead to catastrophic performance and dreadful unfair behavior in the test phase. \citet{ding2021retiring} depicts through extensive experiments that applying post-processing fairness techniques to learn fair predictors of income with respect to race, gender, and age fails to transfer from one US state (training domain) to another state. Several in-processing methods are proposed in the literature to mitigate specific types of distribution shifts. 
\citet{singh2019fair} finds a subset of features with the minimum risk on the domain target with fairness guarantees on the test phase relying on the causal graph and conditional separability of context variables and labels. \citet{du2021fair} proposes a reweighting mechanism~\citep{fang2020rethinking} for learning robust fair predictors when the task involves sample selection bias with respect to different protected groups. 
\citet{rezaei2021robust} generalizes the fair log loss classifier of~\citep{rezaei2020fairness} to a distributionally robust log-likelihood estimator under the covariate shift. \citet{an2022transferring} finds sufficient conditions for transferring fairness from the source to the target domain in the presence of sub-population and domain shifts and handles these two specific types of shifts using a proposed consistency regularizer.  

\vspace{-1mm}

The above methods rely on the availability of unlabeled test samples in the training phase (domain adaptation setting) or explicit assumptions on having access to the causal graph describing the causal interaction of features, and/or knowing the specific type of the distribution shift apriori. As an alternative approach, \citet{taskesen2020distributionally} learns a distributionally robust fair classifier over a Wasserstein ball around the empirical distribution of the training data as the uncertainty set. Unfortunately, their formulation is challenging to solve efficiently via scalable first-order methods, even for linear and convex classifiers such as logistic regression, and there are no efficient algorithms converging to the optimal solutions of the distributional robust problem. The proposed algorithm in the paper is a greedy approach whose time complexity grows with the number of samples. In another effort, \citet{wang2022robust} optimizes the accuracy and maximum mean discrepancy (MMD) of the curvature distribution of the two sub-populations (minorities and majorities) jointly to impose robust fairness. This formulation is based on the idea that flatters local optima with less sharpness have nicer properties in terms of robustness to the distribution shift. However, no convergence guarantee for the proposed optimization problem is provided.
Compared to~\citet{wang2022robust}, our work is amenable to stochastic optimization. In addition, we define the distributionally robust problem \textbf{directly} over the fairness violation measure, while~\citet{wang2022robust} relies on matching the curvature distribution of different sub-populations as a heuristic proxy for the measuring robustness of fairness.  

\vspace{-3mm}
\subsection{Our Contribution} 
\vspace{-2mm}
We propose stochastic and deterministic distributionally robust optimization algorithms under $L_p$-norm balls as the uncertainty sets for maintaining various \textit{group fairness criteria} across different domains. We established the convergence of the algorithms and showed that the group fairness criteria can be maintained in the target domain with the proper choice of the uncertainty set size and fairness regularizer coefficient without relying on \textit{the knowledge of the causal graph of the feature interactions} or explicit knowledge of the distribution shift type (demographic, covariate, or label shift). The proposed stochastic algorithm is the \underline{first provably convergent algorithm with any arbitrary batch size} for distributionally robust fair empirical risk minimization. Stochastic (mini-batch) updates are crucial to large-scale learning tasks consisting of a huge number of data points.

\vspace{-3mm}
\section{Preliminaries and Problem Description}
\vspace{-3mm}
Consider a supervised learning task of predicting a label/target random variable~$y \in \mathcal{Y} \triangleq \{1,2, \ldots, m\}$ based on the input feature $\bx$. Assume our model (e.g. a neural network or logistic regression model) outputs the variable $\hat{y}_{\btheta}(\bx)$  where $\btheta$ is the parameter of the model (e.g. the weights of the neural network). For simplicity, sometimes we use the notation $\hat{y}$ instead of $\hat{y}_{\btheta}(\bx)$. Let $s \in \mathcal{S} = \{1,2,\ldots, k\}$  denote the random variable modeling the sensitive attribute (e.g. gender or ethnicity) of data points. In the fair supervised learning task, we aim to reach two (potentially) competing goals:
Goal 1) Maximize  prediction accuracy~$ \mathbb{P}^*(\hat{y}_{\btheta}(\bx) = y)$;
Goal 2) Maximize fairness.
Here,  $P^*(\cdot)$  is the ground-truth distribution during the deployment/test phase, which is typically unknown during training. The first goal is usually achieved by minimizing a certain loss function. To achieve the second goal, one needs to mathematically define ``fairness'', which is described next.

\vspace{-3mm}
\subsection{Notions of Group Fairness}
\vspace{-1mm}
Different notions of group fairness have been introduced~\citep{zafar2017parity, narayanan2018translation}. Among them, demographic parity~\citep{act1964civil, dwork2012fairness}, equalized odds~\citep{hardt2016equality}, and equality of opportunity~\citep{hardt2016equality} gained popularity.  
A classifier satisfies \textit{demographic parity}~\citep{dwork2012fairness} if the output of the classifier is independent of the sensitive attribute, i.e.,  
\begin{equation}
\label{eq: demographic parity}
    \mathbb{P}^*(\hat{y}, s) = \mathbb{P}^*(\hat{y}) \: \mathbb{P}^*(s)
\end{equation} 
On the other hand, a classifier is fair with respect to the \textit{equalized odds} notion~\citep{hardt2016equality} if
\begin{equation}
\label{eq: equalized_odds}
\small
    \mathbb{P}^*(\hat{y}, s | y = a) = \mathbb{P}^*(\hat{y} | y = a) \: \mathbb{P}^*(s | y = a) \quad \forall a \in \mathcal{Y}.
\end{equation}
Further, in the case of binary classification, \textit{equality of opportunity} is defined as~\citep{hardt2016equality}:
\begin{equation}
\label{eq: equality_of_opportunity}
    \mathbb{P}^*(\hat{y}, s | y = 1) = \mathbb{P}^*(\hat{y} | y = 1) \: \mathbb{P}^*(s | y = 1),
\end{equation}
where $y=1$ is the advantaged group (i.e., the desired outcome from each individual's viewpoint). 

\vspace{-3mm}
\subsection{Fairness Through Regularized ERM}
\vspace{-1mm}
In the fairness notions defined above,
$\hat{y} = \hat{y}_{\btheta} (\bx)$ depends on the parameter of the model $\btheta$. Therefore,  the above fairness notions depend on the parameters of the (classification) model~$\btheta$. Moreover, they are all in the form of (conditional) statistical independence between random variables $\hat{y}, y$, and $s$. 
Thus, a popular framework to reach the two (potentially competing) goals of maximizing accuracy and fairness is through regularized empirical risk minimization framework~\citep{zafar2017fairness, baharlouei2019renyi, mary2019fairness, grari2019fairness, lowy2021fermi}.
In this framework, we add a fairness-imposing regularizer to the regular empirical risk minimization loss. More specifically, the model is trained by solving the optimization problem
\begin{equation}
\label{eq: Fair_Risk_Minimization}
    \min_{\btheta} \quad \mathbb{E}_{\mathbb{P}} \: [\ell( \hat{y}_{\btheta} (\bx), y)] + \lambda \rho \Big(\hat{y},y, s; \mathbb{P}\Big).
\end{equation}
Here, the first term in the objective function aims to improve the model's accuracy with $\ell(\cdot,\cdot)$ being the loss function, such as cross-entropy loss. On the other hand, $\rho (\hat{y}, s, y, \mathbb{P})$ is a group \textit{fairness violation measure}, which quantifies the model's fairness violation. We will discuss examples of such measures in section~\ref{subsec:fairnesViolation}.
$\lambda$ is a positive regularization coefficient to control the tradeoff between fairness and accuracy. In the above formulation, $\mathbb{P}$ denotes the data distribution. Ideally, we would like to define the expectation term~$\mathbb{E}_{\mathbb{P}} [\ell( \hat{y}_{\btheta} (\bx), y)]$ and the fairness violation term~$\rho \Big(\hat{y},y, s; \mathbb{P}\Big)$ in \eqref{eq: Fair_Risk_Minimization} with respect to the test distribution $\mathbb{P}^*$, i.e. , $\mathbb{P} = \mathbb{P}^*$. However, since this distribution is unknown, existing works typically use the training data instead. Next, we describe popular approaches in the literature for defining fairness violation measure~$\rho \Big(\hat{y},y, s; \mathbb{P}\Big)$.

\vspace{-3mm}
\subsection{Measuring Fairness Violation} 
\vspace{-1mm}
\label{subsec:fairnesViolation}
The fairness criteria defined in~\eqref{eq: demographic parity}, \eqref{eq: equalized_odds}, and \eqref{eq: equality_of_opportunity} can all be viewed as a statistical (conditional) independence condition between the random variables $\hat{y},y,$ and $s$. For example, the equalized odds notion~\eqref{eq: equalized_odds} means that $\hat{y}$ is independent of $s$ condition on the random variable~$y$. Therefore, one can quantify the ``violation" of these conditions through well-known statistical dependence measures. Such quantification can be used as a regularization in~\eqref{eq: Fair_Risk_Minimization}. In what follows, we briefly describe some of these measures. \textbf{\small We  only describe these measures and our methodology for the demographic parity notion.} The equalized odds and the equality of opportunity notions can be tackled in a similar fashion.

To quantify/measure fairness violation based on the demographic parity notion, one needs to measure the statistical dependence between the sensitive attribute~$s$ and the output feature~$\hat{y}$. To this end,~\citet{zafar2017fairness} utilizes the empirical covariance of the two variables, i.e., $\rho_c(\hat{y},s, y; \mathbb{P}) \triangleq \mathbb{E} [\hat{y}s]$. However, it is known that if two random variables have $0$ covariance, it does not necessarily imply their statistical independence. To address this issue, several in-processing approaches utilize nonlinear statistical dependence measures such as mutual information~\citep{roh2020fr}, maximum-mean-discrepancy (MMD)~\citep{oneto2020exploiting, prost2019toward},  \Renyi correlation~\citet{baharlouei2019renyi, grari2019fairness}, and Exponential \Renyi Mutual Information (ERMI)~\citep{mary2019fairness, lowy2021fermi}. 
Let us briefly review some of these notions:
Given the joint distribution of sensitive attributes and predictions, the mutual information between the sensitive features and predictions is defined as:
$
\rho_{MI}(\hat{y}, s, y, \mathbb{P}) = \sum_{s, y} \mathbb{P}(s, y) \log \Big( \frac{\mathbb{P}(s, y)}{\mathbb{P}(s) \: \mathbb{P}(y)} \Big).   
$
Some other notions are based on the Hirschfeld-Gebelein-\Renyi (HGR) and Exponential \Renyi Mutual Information (ERMI), which  are respectively defined as~\citep{hirschfeld1935connection, gebelein1941statistische, renyi1959measures, lowy2021fermi}:
\begin{equation}    
    \rho_{\textrm{HGR}}(\hat{y}, s, y, \mathbb{P}) = \sigma_2 (Q)  
    \quad \textrm{and} \quad
    \rho_{\textrm{ERMI}}(\hat{y}, s, y, \mathbb{P}) = {\rm Tr}(Q^T Q). \label{eq:RhoFERMI}
\end{equation}
where $\sigma_2(Q)$ denotes the second largest singular value of matrix $Q$ defined as
\begin{equation} \label{eq:MatrixQ}
    Q_{ij} = \Big[ \frac{\mathbb{P}_{\hat{y}, s}(\hat{y} = i, s = j)}{\sqrt{\mathbb{P}_{\hat{y}}(\hat{y} = i)} \sqrt{\mathbb{P}_{s}(s = j)}} \Big],
\end{equation}

\vspace{-2mm}
\subsubsection{ERMI as Fairness Violation Measure}
\vspace{-1mm}
\label{sec:FairnessViolationThisWork}
In this paper, we utilize $\rho_{\textrm{ ERMI}}(\hat{y}, s, y, \mathbb{P})$ as the measure of independence between the target variable and sensitive attribute(s) due to the following reasons. First, as shown in~\citep{lowy2021fermi}, ERMI upper-bounds many well-known measures of independence such as HGR, $\ell_1$, $\ell_{\infty}$ and Shannon Divergence. Thus, by minimizing ERMI, we also guarantee that other well-known fairness violation measures are minimized. Second, unlike mutual information, \Renyi correlation~\cite{baharlouei2019renyi, mary2019fairness}, and MMD~\citep{prost2019toward} measures, there exists a \textbf{convergent} stochastic algorithms to the stationary solutions of~\eqref{eq: Fair_Risk_Minimization} for ERMI, which makes this measure suitable for large-scale datasets containing a large number of samples~\citep{lowy2021fermi}. 

\vspace{-3mm}
\section{Efficient Robustification of ERMI}
\vspace{-3mm}
In most applications, one cannot access the test set during the training phase. Thus, the regularized ERM framework~\eqref{eq: Fair_Risk_Minimization} is solved w.r.t. the training set distribution. That is, the correlation term~$\rho(\cdot)$ is computed w.r.t. the training data distribution in~\eqref{eq: Fair_Risk_Minimization} to learn~$\btheta$. However, as mentioned earlier, the test distribution may differ from the training distribution, leading to a significant drop in the accuracy of the model and the fairness violation measure~$\rho(\cdot)$ at the test phase.

\vspace{-2mm}
\subsection{Distributionally Robust Formulation of ERMI}
\vspace{-2mm}
A prominent approach to robustify machine learning models against distribution shifts is \textit{Distributionally Robust Optimization} (DRO)~\citep{delage2010distributionally, wiesemann2014distributionally, lin2022distributionally}. In this framework, the performance of the model is optimized with respect to the worst possible distribution (within a certain uncertainty region). Assume that $P_{\textrm{tr}}$ is the joint training distribution of $(\bx, y)$ where $\bx$ is the set of input features, and $y$ is the target variable (label). The distributionally robust optimization for fair risk minimization can be formulated as:
\vspace{-0.05cm}
\begin{equation}
\label{eq: dro_fair_risk_minimization}
\min_{\btheta} \;\; \max_{\mathbb{P} \in \mathcal{U}(\mathbb{P}_{\textrm{tr}})}  \mathbb{E}_{(\bx, y) \sim \mathbb{P}}[\ell(\hat{y}_{\btheta}(\bx), y)]   + \lambda \rho \Big(\hat{y}_{\btheta}(\bx),y, s, \mathbb{P} \Big),
\end{equation}  
where $\mathbb{P}_{\textrm{tr}}$ denotes the empirical distribution over training data points. The set~$\mathcal{U} (\mathbb{P}_{\textrm{tr}})$ is the uncertainty set for test distribution. This set is typically defined as the set of distributions that are close to the training distribution~$\mathbb{P}_{\textrm{tr}}$. The min-max problem in~\eqref{eq: dro_fair_risk_minimization} is non-convex non-concave in general. Therefore, even finding a locally optimal solution to this problem is not guaranteed using efficient first-order methods~\citep{razaviyayn2020nonconvex, daskalakis2021complexity, jing2021towards, pmlr-v202-khalafi23a}. To simplify this problem, we upperbound~\eqref{eq: dro_fair_risk_minimization} by the following problem where the uncertainty sets for the accuracy and fairness terms are decoupled:
\vspace{-0.05cm}
\begin{equation}
\label{eq: dro_upper_bound}
\min_{\btheta} \;\; \max_{\mathbb{P}  \in \mathcal{U}_1(\mathbb{P}_{\textrm{tr}})}  \mathbb{E}_{(\bx, y) \sim \mathbb{P}}[\ell(\hat{y}_{\btheta}(\bx), y)]   + \max_{\mathbb{P}  \in \mathcal{U}_2(\mathbb{P}_{\textrm{tr}})} \lambda \rho \Big(\hat{y}_{\btheta}(\bx),y, s, \mathbb{P} \Big),     
\end{equation}  
The first maximum term of the objective function is the distributionally robust optimization for empirical risk minimization, which has been extensively studied in the literature and the ideas of many existing algorithms can be utilized~\citep{kuhn2019wasserstein, sagawa2019distributionally, levy2020large, zymler2013distributionally}. However, the second term has not been studied before in the literature. Thus, from now on, and to simplify the presentation of the results, we focus on how to deal with the second term. In other words, for now, we consider $\mathcal{U}_1(\mathbb{P}_{\textrm{tr}}) = \{\mathbb{P}_{\textrm{tr}}\}$ is singleton and we only robustify fairness. Later we will  utilize existing algorithms for CVaR and Group DRO optimization  to modify our framework for solving the general form of~\eqref{eq: dro_upper_bound}, and robustify both accuracy and fairness.

\vspace{-3mm}
\subsection{The Uncertainty Set} \label{subsec:UncertaintySet}
\vspace{-1mm}
\normalsize
Different ways of defining/designing the uncertainty set in the DRO framework are discussed in section~\ref{subsec:Related_works}. The uncertainty set can be defined based on the knowledge of the potential type of distribution shifts (which might not be available during training). Moreover, the uncertainty set should be defined in a way that the resulting DRO problem is efficiently solvable. 
As mentioned in section~\ref{sec:FairnessViolationThisWork}, we utilize the correlation measure $\rho_{\text{ERMI}}(\cdot)$, which depends on the probability distribution $\mathbb{P}$ through matrix~$Q$; see equations~\eqref{eq:MatrixQ} and~\eqref{eq:RhoFERMI}. Consequently, we define the uncertainty set directly on the matrix~$Q$.  Thus, our robust fair learning problem can be expressed as
\normalsize
\begin{equation}\tag{Dr. FERMI}
\label{eq: fermi_fair_empirical_risk}
        \min_{\btheta} \;\;  \mathbb{E}_{\mathbb{P}_{\textrm{tr}}}[\ell(\hat{y}_{\btheta}(\bx), y)]   +\lambda \max_{Q_{\btheta} \in \mathcal{B} (Q_{\btheta}^{\textrm{tr}},\epsilon)} {\rm Tr}(Q_{\btheta}^T Q_{\btheta}),   
\end{equation}
\normalsize
where $Q_{\btheta}^{\textrm{tr}}$ is obtained by plugging $\mathbb{P}_{\textrm{tr}}$ in~\eqref{eq:MatrixQ}, and $\mathcal{B} (Q_{\btheta}^{\textrm{tr}},\epsilon)$ is the uncertainty set/ball around $Q_{\btheta}^{\textrm{tr}}$.
\normalsize

\vspace{-2mm}
\subsection{Simplification of Dr. FERMI}
\vspace{-1mm}
The formulation~\eqref{eq: fermi_fair_empirical_risk} is of the min-max form and could be challenging to solve at first glance~\cite{razaviyayn2020nonconvex, daskalakis2021complexity}. The following theorem shows how \eqref{eq: fermi_fair_empirical_risk} can be simplified for various choices of the uncertainty set~$\mathcal{B} (Q_{\btheta}^{\textrm{tr}},\epsilon)$, leading to the development of efficient optimization algorithms. The proof of this theorem is deferred  to Appendix~\ref{appendix: A}
\vspace{-3mm}
\begin{theorem} 
\label{thm: robust_L}
Define $\normalfont \mathcal{B}(Q_{\btheta}^{\textrm{tr}}, \epsilon, p) = \{ Q: \|\sigma(Q) - \sigma(Q_{\btheta}^{\textrm{tr}})\|_p \leq \epsilon\}$ where $\sigma(Q)$ is the vector of singular values of $Q$. Then: 

\noindent \textbf{a)} for $p = 1$, we have:
\vspace{-2mm}
\begin{equation}
    \normalfont
    \max_{Q_{\btheta} \in \mathcal{B} (Q_{\btheta}^{\textrm{tr}},\epsilon,p), \sigma_1(Q_{\btheta})=1} {\rm Tr}(Q_{\btheta}^T Q_{\btheta}) = {\rm Tr}\Big((Q_{\btheta}^{\textrm{tr}})^T Q_{\btheta}^{\textrm{tr}}\Big) + 2 \epsilon \sigma_2 \Big(Q_{\btheta}^{\textrm{tr}}\Big) + \epsilon^2, 
    \label{eq:robust_l1}
\end{equation}
which means the robust fair regularizer can be described as $ \rho_{\textrm{ERMI}}(\hat{y}, s) + 2\epsilon \rho_{\textrm{HGR}}(\hat{y}, s)$  when $p=1$.

\noindent \textbf{b)} For $p = 2$,
\vspace{-4mm}
\begin{equation}
\normalfont
    \max_{Q_{\btheta} \in \mathcal{B} (Q_{\btheta}^{\textrm{tr}},\epsilon,p)} {\rm Tr}(Q_{\btheta}^T Q_{\btheta}) = {\rm Tr}\Big((Q_{\btheta}^{\textrm{tr}})^T Q_{\btheta}^{\textrm{tr}}\Big) + 2\epsilon \sqrt{{\rm Tr} \Big((Q_{\btheta}^{\textrm{tr}})^T Q_{\btheta}^{\textrm{tr}}\Big)} + \epsilon^2,
\label{eq:Robust_FERMI_L2}
\end{equation}
which means that when $p=2$, the regularizer is equivalent to $\rho_{\textrm{ ERMI}}(\hat{y}, s) + 2\epsilon \sqrt{\rho_{\textrm{ ERMI}}(\hat{y}, s)}$.

\noindent \textbf{c)} For $p = \infty$, assume that $Q_{\btheta}^{\textrm{tr}} = U_{\btheta} \Sigma_{\btheta} V_{\btheta}^T$ is the singular value decomposition of $Q_{\btheta}^{\textrm{tr}}$. Therefore: 
\vspace{-2mm}
\begin{equation} 
\normalfont
    \max_{Q_{\btheta} \in \mathcal{B} (Q_{\btheta}^{\textrm{tr}},\epsilon,p)} {\rm Tr}(Q_{\btheta}^T Q_{\btheta}) =  {\rm Tr}\Big((Q_{\btheta}^{\textrm{tr}})^T Q_{\btheta}^{\textrm{tr}}\Big) + 2\epsilon \Tr(|\Sigma_{\btheta}|) + \epsilon^2.
\label{eq:Robust_FERMI_L_inf}
\end{equation}
\end{theorem}

\begin{remark}
    In the case of $p=1$, we enforced~$\sigma_1(Q) = 1$. Without this condition, as we will see in the proof, the maximum is attained for a $Q$ which does not correspond to a probability vector.
\end{remark}

\vspace{-2mm}
\subsection{Generalization of Dr. FERMI}
\vspace{-2mm}
A  natural question on the effectiveness of the proposed DRO formulation for fair empirical risk minimization is whether we can guarantee the boundedness of the fairness violation by optimizing~\eqref{eq: fermi_fair_empirical_risk}. The following theorem shows that if $\lambda$ and $\epsilon$ are appropriately chosen, then the fairness violation of the solution of~\eqref{eq: fermi_fair_empirical_risk} can be reduced to any small positive value on the unseen test data for any distribution shift. This result is proven for the cross-entropy loss (covering logistic regression and neural networks with cross-entropy loss). Following the proof steps, the theorem can be extended to other loss functions that are bounded below (such as hinge loss and mean squared loss). 
\vspace{-2mm}
\begin{theorem}
\label{thm: fairness_generalization}
Let $\ell(\cdot, \cdot)$ in~\eqref{eq: fermi_fair_empirical_risk} be the cross-entropy loss. For any shift in the distribution of data from the source to the target domain, and for any given $\gamma > 0$, there exists a pair of $(\lambda, \epsilon)$ such that the solution to~\eqref{eq: fermi_fair_empirical_risk} has a demographic parity violation less than or equal to $\gamma$ on the target domain (unseen test data).    
\end{theorem}
\vspace{-2mm}
The proof is deferred to Appendix~\ref{appendix: generalization}.
A discussion on the significance and limitations of the theorem can be found in Appendix~\ref{appendix: generalization_limitations}. 

\vspace{-3mm}
\section{Algorithms for Solving~\eqref{eq: fermi_fair_empirical_risk}}
\label{sec: algorithms}
\vspace{-3mm}
In this section, we utilize Theorem~\ref{thm: robust_L} to first develop efficient deterministic algorithms for solving~\eqref{eq: fermi_fair_empirical_risk}. Building on that, we will then develop stochastic (mini-batch) algorithms. 

\vspace{-2mm}
\subsection{Deterministic (Full-Batch)  Algorithm}
\vspace{-2mm}
Let us first start by developing an algorithm for the case of $p=1$, as in~\eqref{eq:robust_l1}. Notice that
\begin{equation}
\label{Min-Max-Fairness_DP-discrete}
\sigma_2 (Q_{\btheta}^T Q_{\btheta}) = \max_{\bv \perp \bv_1, \, \|\bv\|^2 \leq 1} \sqrt{\bv^T Q_{\btheta}^T Q_{\btheta}\bv}.
\end{equation}
where $\bv_1 = \Big[\sqrt{\mathbb{P} (S = s_1)}, \ldots, \sqrt{\mathbb{P} (S = s_1)}\Big]$; see also \citep[Equation (6)]{baharlouei2019renyi}.
Further, as described in~\citep[Proposition 1]{lowy2021fermi}, we have 
\vspace{-1mm}
\begin{equation} \label{eq:ERMIestimation}
\small
\widehat{\rho}_{\textrm{ERMI}}(\hat{y}, s, y; \mathbb{P}) = \max_{W}
\{-\Tr(W \widehat{P}_{\hat{y}} W^T) 
+2 \Tr(W \widehat{P}_{\hat{y}, s} \hat{P}_{s}^{-1/2}) - 1 
\}
\end{equation}
\normalsize
where $\widehat{\rho}_{\textrm{ERMI}}$ is the ERMI correlation measure~\eqref{eq:RhoFERMI} defined on training distribution~$\mathbb{P}_{{\rm tr}}$. Here, $\widehat{P}_{\hat{y}, s}$ is a probability matrix whose $(i, j)$-th entry equals $\mathbb{P}_{\textrm{tr}}(\hat{y} = i, s = j)$. 
Similarly, we  define $\widehat{P}_{\hat{y}}$.

Combining~\eqref{eq:robust_l1} with \eqref{Min-Max-Fairness_DP-discrete} and \eqref{eq:ERMIestimation} leads to a min-max reformulation of~\eqref{eq: fermi_fair_empirical_risk} in variables~$(W,\bv,\btheta)$ (see Appendix~\ref{appendix: gradient_computation} for details). This reformulation gives birth to Algorithm~\ref{alg: robust_l1}. At each iteration of this algorithm, we maximize w.r.t. $\bv$ and $W$ variables, followed by a step of gradient descent w.r.t. $\btheta$ variable. One can establish the convergence of this algorithm following standard approaches in the literature; see Theorem~\ref{thm: Fair-classification-} in Appendix~\ref{appendix:ConvergenceofDeterministic}. Note that $\hat{P}_{\hat{y}}$ and $\hat{P}_{\hat{y}, s}$ are functions of $\btheta$ (through $\hat{y}$). Thus, it is crucial to recompute them after each update of $\btheta$. However, as $\widehat{P}_{s}$ does not depend on $\theta$, it remains the same throughout the training procedure. 
\begin{algorithm}
    \caption{Deterministic Distributionally Robust FERMI under $L_1$ Ball Uncertainty}
    \label{alg: robust_l1}
    \begin{algorithmic}[1]
	 \STATE \textbf{Input}: $\btheta^0 \in \mathbb{R}^{d_{\theta}}, ~W^0 = 0$, step-sizes $\eta$, fairness parameter $\lambda \geq 0,$ iteration number $T$.
    
    \FOR {$t = 1, \ldots, T$}
    
    \STATE \small $\btheta^{t} = \btheta^{t-1} - \eta \Big(\frac{1}{n}\sum_{i=1}^n \nabla_{\btheta} \ell(\hat{y}_{\btheta^{t-1}}(\bx_i), y) + \lambda \epsilon \nabla_{\btheta}\Big(\sqrt{\bv^T Q_{\btheta^{t-1}}^T Q_{\btheta^{t-1}}\bv}\Big)  + \lambda \Tr \big(\nabla_{\btheta}(Q_{\btheta^{t-1}}^T Q_{\btheta^{t-1}})\big)\Big) $ 
    \vspace{1mm}

    \STATE Set \small $W^{t} = 
    \widehat{P}_{s}^{-1/2} \widehat{P}_{\hat{y}, s}^T \widehat{P}_{\hat{y}}^{-1}$
    \vspace{1mm}

    \STATE Set $\bv^{t}$ to the second largest singular vector of $Q_{\btheta^t}$ by performing SVD on $Q_{\btheta^t}$.
    \vspace{1mm}
    
    \STATE Update $\widehat{P}_{\hat{y}}$ and $\widehat{P}_{\hat{y}, s}$ as a function of $\btheta_{t-1}$. 
    \ENDFOR
    \STATE \textbf{Return:} $\btheta^T, W^T, \bv^T.$
\end{algorithmic}
\vspace{-.03in}
\end{algorithm}
Details on the gradient computations in Algorithm~\ref{alg: robust_l1} are deferred to Appendix~\ref{appendix: gradient_computation}. 
Following similar steps, we developed  algorithms for $L_2$ and $L_{\infty}$ uncertainty balls using~\eqref{eq:Robust_FERMI_L2} and \eqref{eq:Robust_FERMI_L_inf}; see Appendix~\ref{app: D}. 

\vspace{-2mm}
\subsection{Stochastic (Mini-Batch) Algorithm}
\label{sec: stochastic}
\vspace{-2mm}
In large-scale learning problems, we can only use small batches of data points to update the parameters at each iteration. Thus, it is crucial to have stochastic/minibatch algorithms. The convergence of such algorithms relies on the unbiasedness of the (minibatch) gradient at each iteration. To develop a stochastic algorithm version of Dr. FERMI, one may na\"ively follow the steps of Algorithm~\ref{alg: robust_l1} (or a gradient descent-ascent version of it) via using mini-batches of data. However, this heuristic does not converge (and also leads to statistically biased measures of fairness). It is known that for the stochastic gradient descent algorithm to converge, the update direction must be a statistically unbiased estimator of the actual gradient of the objective function~\citep{polyak1990new, nemirovski2009robust}; thus, the na\"ive heuristic fails to converge and may even lead to unfair trained models. The following lemma rewrites the objective in~\eqref{eq:Robust_FERMI_L2} so that it becomes a summation/expectation over the training data and hence provides a statistically unbiased estimator of the gradient for stochastic first-order algorithms. 
\vspace{-2mm}
\begin{lemma}
\label{lemma:stochastic_reformulation}
Let $\normalfont \mathcal{B}(Q_{\btheta}^{\textrm{tr}}, \epsilon) = \{ Q: \|Q - Q_{\btheta}^{\textrm{tr}}\|_2 \leq \epsilon\}$,  then~\eqref{eq: fermi_fair_empirical_risk} is equivalent to:
\vspace{-1mm}
\begin{equation}
\label{eq:robust_l2_reformulation}
    \min_{\alpha>0, \boldsymbol{\theta}} \: \max_{W \in \mathbb{R}^{k\times d}} \frac{1}{n} \big[ \sum_{i=1}^{n} \ell(\bz_i; \btheta) + \lambda(1 + \epsilon \alpha) \psi(\bz_i; \btheta, W) \big] + \frac{\lambda \epsilon}{\alpha}
\end{equation}
Where $\psi$ is a quadratic concave function in $W$ defined as:
\begin{equation}
\label{eq: psi}
\small
\normalfont
  \psi(\bz_i; \btheta, W) :=  -\Tr(W \expec[\widehat{\by}_{\btheta}(\bx_i) \widehat{\by}_{\btheta}(\bx_i)^T | \bx_i] 
  W^T) + 2 \Tr(W \expec[\widehat{\by}_{\btheta}(\bx_i) \bs_i^T | \bx_i, s_i] \widehat{P}_{s}^{-1/2})
\end{equation}
\end{lemma} 
Lemma~\ref{lemma:stochastic_reformulation} is key to the development of our convergent stochastic algorithms (proof in Appendix~\ref{appendix: B}). It rewrites~\eqref{eq: fermi_fair_empirical_risk} as the average over $n$ training data points by introducing new optimization variables $\alpha$ and  $W$. Consequently, taking the gradient of the new objective function w.r.t. a randomly drawn batch of data points is an unbiased estimator of the full batch gradient. Having an unbiased estimator of the gradient, we can apply the Stochastic Gradient Descent Ascent (SGDA) algorithm to solve~\eqref{eq:robust_l2_reformulation}. The details of the algorithm and  its convergence proof can be found in Appendix~\ref{appendix: stochastic}.

\vspace{-3mm}
\section{Robustification of the Model Accuracy}
\vspace{-3mm}
In Section~\ref{sec: algorithms}, we analyzed the DRO formulation  where $\mathcal{U}_1(\mathbb{P}_{\textrm{tr}})$ is singleton in~\eqref{eq: dro_upper_bound}. However, to make the accuracy of the model robust, we can consider various choices for $\mathcal{U}_1(\mathbb{P}_{\textrm{tr}})$. For example, we can consider the set of distributions with the bounded likelihood ratio to $\mathbb{P}_{\textrm{tr}}$, which leads to Conditional Value at Risk (CVaR)~\citep{rockafellar2000optimization}. Using the dual reformulation of CVaR, \eqref{eq: dro_upper_bound} can be simplified to (see~\citep[Appendix A]{levy2020large} and \citep[Chapter 6]{shapiro2021lectures}):
\vspace{-1mm}
\begin{equation}
\label{eq: dro_cvar}
\tag{CVaR-Dr. FERMI}
\min_{\btheta, \eta > 0} \; \frac{1}{\alpha} \mathbb{E}_{(\bx, y) \sim \mathbb{P}_{\textrm{tr}}}\Big[\ell \big(\hat{y}_{\btheta}(\bx), y \big) - \eta \Big]_{+} + \eta + \max_{\mathbb{P}  \in \mathcal{U}_2(\mathbb{P}_{\textrm{tr}})} \lambda \rho \Big(\hat{y}_{\btheta}(\bx),y, s, \mathbb{P} \Big),     
\end{equation}
where $[x]_{+}$ is the projection to the set of non-negative numbers and we define $\mathcal{U}_2(\mathbb{P}_{\textrm{tr}})$ the same way as we defined in subsection~\ref{subsec:UncertaintySet}. All methods and algorithms in Section~\ref{sec: algorithms} can be applied to~\eqref{eq: dro_cvar}. Compared to the ERM, the CVaR formulation has one more minimization parameter $\eta$ that can be updated alongside $\btheta$ by the (stochastic) gradient descent algorithm. Similarly, one can use group DRO formulation for robustifying the accuracy~\citep{sagawa2019distributionally}. Assume that $\mathcal{G}$ contains the groups of data (in the experiments, we consider each sensitive attribute level as one group). Then,~\eqref{eq: dro_upper_bound} for group DRO as the measure of accuracy can be written as:
\vspace{-1mm}
\begin{equation}
\label{eq: dro_group}
\tag{Group-Dr. FERMI}
\min_{\btheta} \;\; \max_{g \in \mathcal{G}} \; \mathbb{E}_{(\bx, y) \sim P_g}[\ell(\hat{y}_{\btheta}(\bx), y)]   + \max_{\mathbb{P}  \in \mathcal{U}_2(\mathbb{P}_{\textrm{tr}})} \lambda \rho \Big(\hat{y}_{\btheta}(\bx),y, s, \mathbb{P} \Big),     
\end{equation}
Different variations of Algorithm~\ref{alg: robust_l1} for optimizing~\eqref{eq: dro_cvar} and~\eqref{eq: dro_group} are presented in Appendix~\ref{appendix: cvar_alg}. These developed algorithms are evaluated in the next section.

\vspace{-3mm}
\section{Numerical Experiments}
\label{sec: numerical_experiments}
\vspace{-3mm}
We   evaluate the effectiveness of our proposed methods in several experiments. 
We use two well-known group fairness measures: Demographic Parity Violation (DPV) and Equality of Opportunity Violation (EOV)~\citep{hardt2016equality}, defined as \normalsize
{\small
\begin{equation*}
    \textrm{DPV} = | P(\hat{y} = 1 | s = 1) - P(\hat{y} = 1 | s = 0) |,\; \textrm{and}\;
    \textrm{EOV} = |P(\hat{y} = 1 | s = 1, y=1) - P(\hat{y} = 1 | s = 0, y=1)|
\end{equation*}
}
The details of our hyperparameter tuning are provided in Appendix~\ref{app:HyperParamCode}. All implementations for experiments are available at \url{https://github.com/optimization-for-data-driven-science/DR-FERMI}. More plots and experiments are available in Appendix~\ref{app:AdditionalExperiments} and Appendix~\ref{app: stochastic_plot}.
\normalsize
\vspace{-2mm}
\subsection{Modifying Benchmark Datasets to Contain Hybrid Distribution Shifts}
\begin{figure}[H]
\vspace{-3mm}
    \begin{center}
    \centerline{\includegraphics[width=0.99\columnwidth]{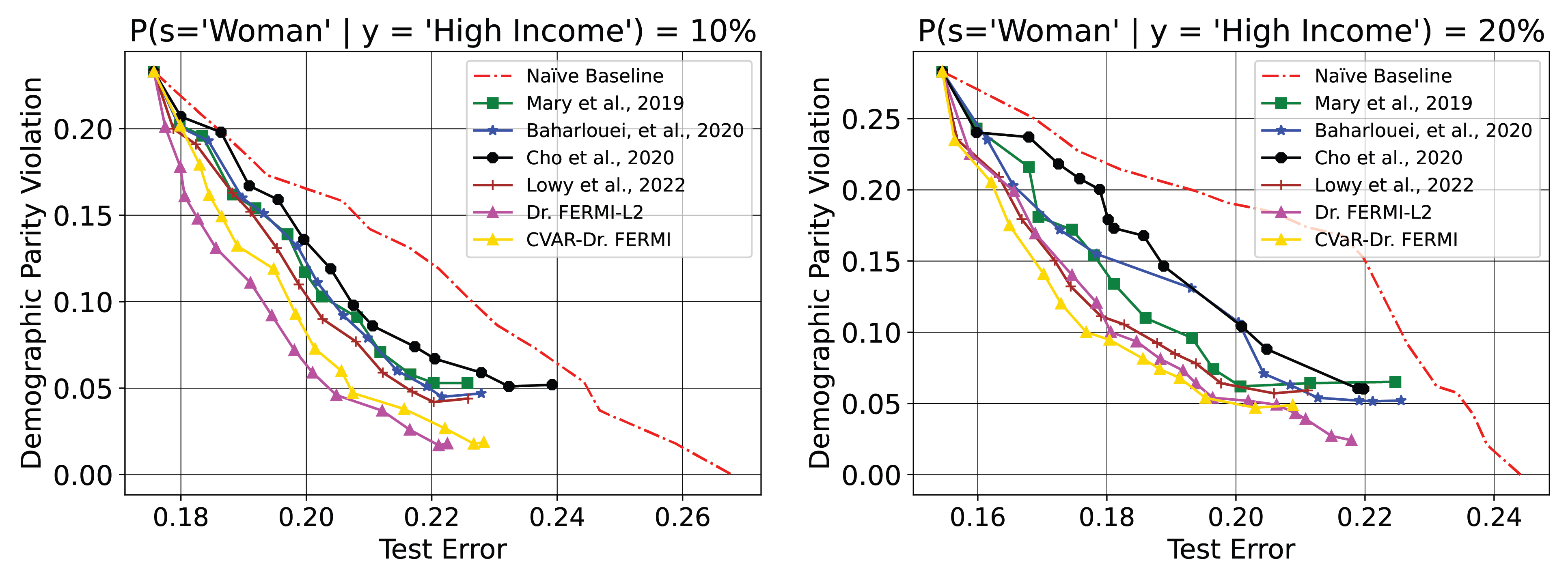}}
    \vspace{-4mm}
    \caption{\small Performance of the trained fair models on test datasets that have different distributions than the training data.  In the left figure, we undersampled the high-income minority group (women), while we oversampled in the right figure. The proposed methods (Dr. FERMI and CVaR-Dr. FERMI) generally outperform other benchmarks. In both figures, either Dr. FERMI or CVaR-Dr. FERMI can reach lower demographic parity violations, given any accuracy level. The red dashed line represents the Na\"ive baseline where the model outputs zero with probability $p$. By increasing $p$, the model becomes fairer at the cost of the loss in accuracy.}
    \label{fig: adult}
\end{center}
\vspace{-10mm}
\end{figure}

\vspace{-2mm}
Standard benchmark datasets in algorithmic fairness, such as Adult, German Credit, and COMPAS~\citep{Dua:2019}, include test and training data that follow the same distribution. 
To impose distribution shifts, we choose a subset of the test data to generate datasets containing demographic and label shifts.
The demographic shift is the change in the population of different sensitive sub-populations from the source and the target distributions, i.e., $\mathbb{\widehat{P}}_s (s) \neq \mathbb{P}^*_s (s)$. Similarly, the label shift means $\mathbb{\widehat{P}}_y (y) \neq \mathbb{P}^*_y (y)$. 
The Adult training (and test) data has the following characteristics:
\vspace{-1mm}
\begin{equation*}
    \mathbb{\widehat{P}}_s (s = \textrm{`Woman'}) = 33.07\% \quad \textrm{and} \quad 
    \mathbb{\widehat{P}}_s (s = \textrm{`Woman'} | \: y = \textrm{`High Income'}) = 15.03\%
\end{equation*}
We generate two different test datasets containing distribution shifts where the probability $\mathbb{\widehat{P}}_s (s = \textrm{`Woman'} | y = \textrm{`High Income'})$ is $10\%$ and $20\%$  (by undersampling and oversampling high-income women). We train the model on the original dataset and evaluate the performance and fairness (in terms of demographic parity violation) on two newly generated datasets in Figure~\ref{fig: adult}.
\vspace{-0mm}
\subsection{Robustness Against Distribution Shift in Real Datasets}
\vspace{-1mm}
While the original Adult dataset~\citep{Dua:2019} has no considerable distribution shift, a relatively new dataset ACS-Income~\citep{ding2021retiring} designed based on the US census records contains natural distribution shifts since we can choose different US states as source and target datasets. The underlying task defined on the dataset is to predict whether a given person is high-income or low-income.
The sensitive attributes are gender and race. 
Using this data, we perform experiments on binary and non-binary sensitive attribute cases. In the binary setting, we only consider gender as the sensitive attribute (in the dataset there are only two genders). In the non-binary case, we have four different sensitive groups: White-Male, White-Female, Black-Male, and Black-Female (a combination of race and gender). One can observe that  $P_s(s)$ and $P_{y, s} (y, s)$ have large fluctuations over different states. Thus, these datasets, as mentioned in the paper~\citep{ding2021retiring}, naturally contain the distribution shift with respect to different states. 

\begin{figure}[h]
    \begin{center}
    \vspace{-3mm}
    \centerline{\includegraphics[width=0.99\columnwidth]{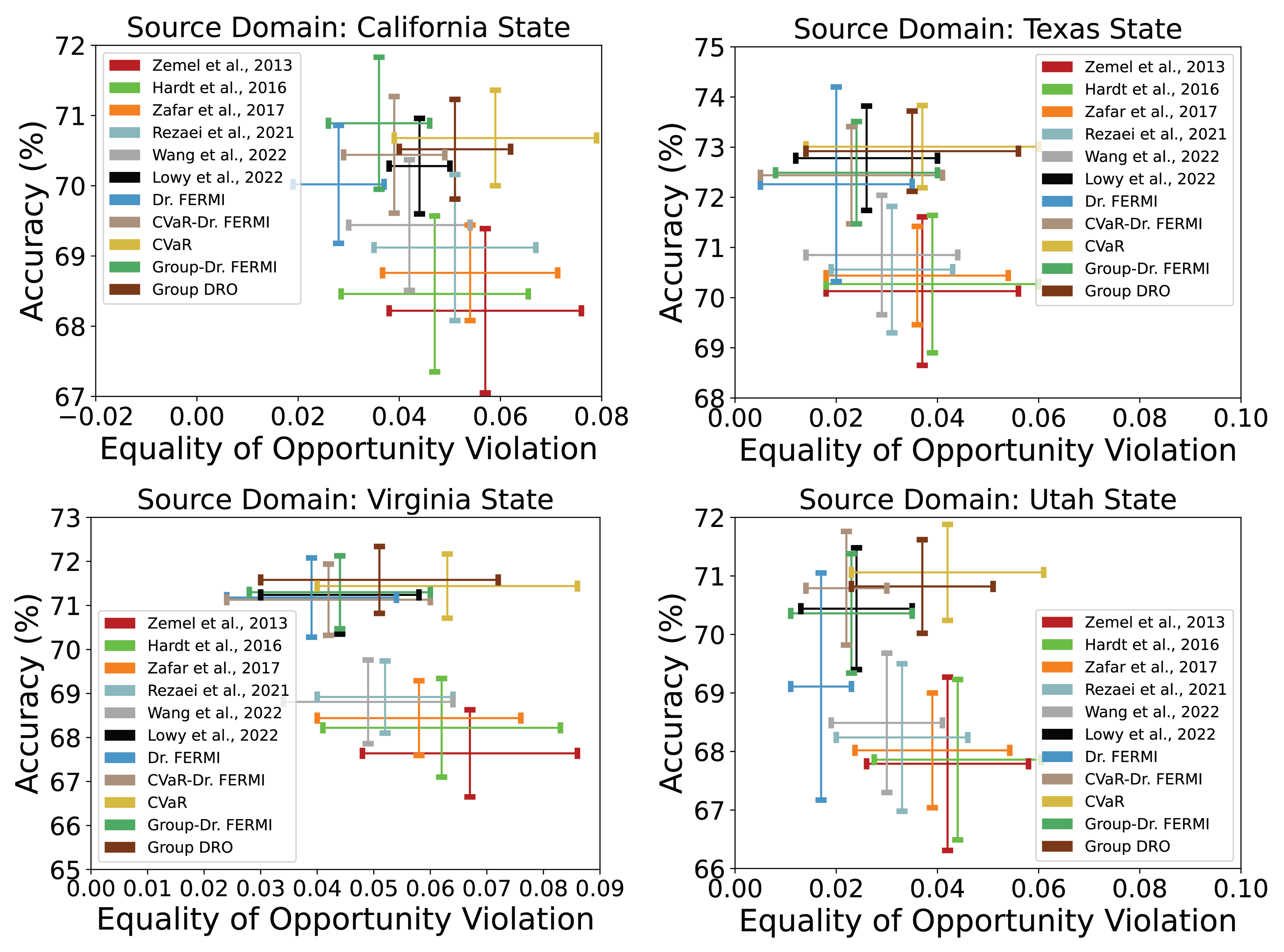}}
    \vspace{-3mm}
    \caption{\small Learning fair logistic regression models on four different states as the source (in-distribution) domain and evaluating them on all $50$ states of the US.}
    \label{fig: acs}
\end{center}
\end{figure}

In Figure~\ref{fig: acs}, we apply  Hardt Post-Processing approach~\citep{hardt2016equality}, Zemel Pre-processing method~\citep{zemel2013learning}, FERMI~\citep{lowy2021fermi}, Robust Log-Loss under covariate shift~\citep{rezaei2021robust}, Curvature matching with MMD~\citep{wang2022robust}, and our distributionally robust method under $L_2$ norm for three accuracy variations (Dr. FERMI, CVaR-Dr. FERMI, and Group-Dr. FERMI), on the new adult (ACS-Income) dataset~\citep{ding2021retiring}. The methods are trained on a single state (California, Texas, Utah, and Virginia) and evaluated/tested on all $50$ states in terms of prediction accuracy and fairness violation under the equality of opportunity notion. We chose California and Texas in accordance with other papers in the literature as two datasets with a large number of samples. Further, we chose Utah and Virginia as the two states with the highest and lowest initial equality of opportunity violations. 

Each method's horizontal and vertical range in Figure~\ref{fig: acs} denote $25$-th and $75$-th percentiles of accuracy and fairness violations across $50$ states, respectively. Thus, if a line is wider, it is \textit{less robust} to the distribution shift. Ideally, we prefer models whose corresponding curves are on the upper-left side of the plot. Figure~\ref{fig: acs} shows that Dr. FERMI with $L_2$ uncertainty ball has a better robust, fair performance than other approaches among $50$ states when the training (in-distribution) dataset is either of the four aforementioned states. When maintaining more accuracy is a priority, one can use CVaR or Group DRO as the objective function for the accuracy part instead of the ERM. As a note, we can see that learning the model on a dataset with a more initial fairness gap (Utah) leads to a better generalization in terms of fairness violation, while learning on datasets with a less fairness gap (Virginia) leads to poorer fairness in the test phase.
\vspace{-3mm}
\subsection{Handling Distribution Shift in Non-Binary Case}
\vspace{-2mm}
We run different versions of Dr. FERMI alongside~\citet{mary2019fairness}, \citet{baharlouei2019renyi}, \citet{cho2020fair}, and~\citet{lowy2021fermi} as the baselines supporting multiple sensitive attributes. The algorithms are executed on the ACS PUMS~\citep{ding2021retiring} dataset with gender and race as the sensitive attributes. The accuracy and equality of opportunity violations are calculated on $50$ different datasets, and the average is reported in Table~\ref{tab: non_binary}. Training is done on California data. Dr. FERMI under $L_2$ ball has the best performance in terms of average fairness violation across $50$ states. For robust accuracy, we suggest using CVaR for robustifying the accuracy. Note that, in all cases, the training equality of opportunity violation is set to \textbf{0.02} for all methods in the table.

\noindent \textbf{Evaluation of Stochastic DR ERMI: }
 batch sizes. By reducing the batch size from full-batch to small batches, we observed that our algorithm's performance remains nearly the same, while other benchmarks' performance varies significantly. Further, we report the time and memory consumption of Dr. FERMI and other benchmarks on the experiment presented in Table~\ref{tab: non_binary}. The results of these experiments are available in Appendix~\ref{app: stochastic_plot}. 

\begin{table*}
\label{tab: non_binary}
\vspace{-8mm}
\centering
\begin{tabular}{|c|c|c|c|}
\hline
\textbf{Method}         & \textbf{Tr Accuracy} & \textbf{Test Accuracy} & \textbf{Test EO Violation} \\ \hline
Mary et al., 2019 & 71.41\%  & 68.35\% & 0.1132 \\ \hline
Cho et al., 2020 & 71.84\% & 68.91\% & 0.1347 \\ \hline
Baharlouei et al., 2020 & 72.77\% & 69.44\% & 0.0652                 \\ \hline
Lowy et al., 2022 & 73.81\% & 70.22\% & 0.0512 \\ \hline
\textbf{Dr. FERMI-$L_1$} & 73.45\%  & 70.09\% & 0.0392 \\ \hline
\textbf{Dr. FERMI-$L_2$} & 73.12\% & 69.71\% & \textbf{0.0346}         \\ \hline
\textbf{Dr. FERMI-$L_{\infty}$} & 73.57\% & 69.88\% & {\ul 0.0359}                     \\ \hline
\textbf{CVaR-Dr. FERMI-$L_1$} & \textbf{74.21}\%  &  \textbf{70.94\%} & 0.0471 \\ \hline
\textbf{CVaR-Dr. FERMI-$L_2$} & 73.84\% & 70.26\% & 0.0429 \\ \hline
\textbf{CVaR-Dr. FERMI-$L_{\infty}$} & {\ul 73.92\%} & {\ul 70.45\%}  & 0.0466 \\ \hline
\end{tabular}
\vspace{-1mm}
\caption{\small Train and Test Accuracy and Fairness Violation of Different Methods on ACS PUBS dataset. Dr. FERMI under $L_2$ ball achieves the best fairness violation on average among different states. The distributionally robust algorithm under $L_1$ ball has a better accuracy but is less fair than $L_2$.}
\end{table*}

\newpage
\bibliographystyle{abbrvnat}
\bibliography{references}

\newpage
\appendix
\newpage
\appendix

\section{Proof of Theorem~\ref{thm: robust_L}}
\label{appendix: A}
Let~$Q_{\btheta} = U \Sigma V$ and $Q_{\btheta}^{\textrm{tr}} = U_0 \Sigma_0 V_0$ be the singular value decompositions of matrices $Q_{\btheta}$ and $Q_{\btheta}^{\textrm{tr}}$ respectively. We aim to solve the following optimization problem:
\begin{equation}
\label{eq: lp_problem}
    \max_{Q_{\btheta}}\;  \Tr(Q^T Q) \quad \textrm{s.t. } \: \| \Sigma - \Sigma_{0}\|_p \leq \epsilon.
\end{equation}
Since $U$ and $V$ are unitary matrices ($U^T U = V^T V = I$), we have $\Tr(Q^T Q) = \Tr(\Sigma^T\Sigma)$. 
Therefore, Problem~\eqref{eq: lp_problem} is equivalent to:
\begin{equation}
\label{eq: lp_problem_eq}
    \max_{Q} \;  \Tr(\Sigma^T \Sigma) \quad \textrm{s.t. } \: \| \Sigma - \Sigma_{\btheta}\|_p \leq \epsilon.
\end{equation}

\noindent \textbf{a)} If $p = 1$, we aim to solve the following maximization problem:
\begin{equation}
\label{eq: l1_problem}
    \max_{\Sigma} \;  \Tr(\Sigma^T \Sigma) \quad \textrm{s.t. } \: \| \Sigma - \Sigma_{0}\|_1 \leq \epsilon.
\end{equation}
Let $\ba$ and $\bb$ be the vectorized forms of $\Sigma$ and $\Sigma_0$ respectively. Therefore, the above problem is equivalent to the following:
\begin{equation}
\label{eq: l1_problem_vec}
    \max_{\ba} \; \sum \ba_i^2 \quad \textrm{s.t. } \: \| \ba - \bb\|_1 \leq \epsilon
\end{equation}
Without loss of generality, assume that $|\bb_1| \geq \dots \geq |\bb_m|$. We claim the solution to this problem is obtained by putting all budgets to the largest entry of $\ba$ ($\ba_1^* = \bb_{1} + \epsilon$, and $\ba_i^* = \bb_{i} \quad \forall i > 1$). To arrive at a contradiction, assume that there exists an optimal solution such that $\ba_1 = \bb_{1} + \epsilon_1$ and $\ba_2 = \bb_{2} + \epsilon_2$ such that $\epsilon_2 > 0$. We claim a solution with $\ba^{'}_1 = \bb_{1} + \epsilon_1 + \epsilon_2$ and $\ba^{'}_2 = \bb_{2}$ has a higher objective value function:
\begin{gather*}
    \ba^{'2}_1 + \ba^{'2}_2 = (\bb_{1} + \epsilon_1 + \epsilon_2)^2 + \bb_{2}^2 \\ = \bb_{1}^2 + \bb_{2}^2 + \epsilon_1^2 + \epsilon_2^2 + 2\epsilon_1 \epsilon_2 + 2\epsilon_1 \bb_1 + 2\epsilon_2 \bb_1 \\  > \bb^{2}_{1} + \epsilon_1^2 + 2\epsilon_1 \bb_1 + \bb^{2}_{2}  + \epsilon_2^2 + 2\epsilon_2 \bb_1 \\ \geq \bb^{2}_{1} + \epsilon_1^2 + 2\epsilon_1 \bb_1 + \bb^{2}_{2}  + \epsilon^2 + 2\epsilon_2 \bb_2 \\ = (\bb_1 + \epsilon_1)^2 + (\bb_2 + \epsilon_2)^2  = \ba_1^2 + \ba_2^2
\end{gather*}
Thus, with proof by contradiction, we have $\epsilon_2 = 0$. 
Similarly for other $\ba_i$'s such that $i > 2$, if the optimal $\ba_i = \bb_i + \epsilon_i$, then $\epsilon_i = 0$. Hence, the claim that the optimal solution to Problem~\eqref{eq: l1_problem_vec} is obtained by putting $\epsilon$ to the largest entry is true. Note that the largest singular value of $Q_{\btheta}$ is always $1$. We also consider the same constraint on matrix $Q$ (this is a practical assumption since the actual matrix $Q$ generated by $P^{*}$, the ground-truth joint distribution of data in the test domain, has the largest singular value equal 1). Thus, to maximize the summation of singular values, we put all available budgets to the second-largest singular value. Therefore, the optimal solution to Problem~\eqref{eq: l1_problem} is $\Tr(Q_{\btheta}^T Q_{\btheta}) + 2\epsilon \sigma_2 (Q_{\btheta}) + \epsilon^2$, which is the desired result in Theorem~\ref{thm: robust_L} part \textbf{a}.

\noindent \textbf{b)} If $p=2$, we require to solve the following maximization problem:
\begin{equation}
\label{eq: l2_problem_eq}
    \max_{\Sigma} \Tr(\Sigma^T \Sigma) \quad \textrm{s.t. } \: \| \Sigma - \Sigma_{0}\|_2 \leq \epsilon,
\end{equation}

Assume that $\ba = \textrm{diag}(\Sigma)$ and $\bb = \textrm{diag}(\Sigma_0)$ 
Thus, Problem~\eqref{eq: l2_problem_eq} is equivalent to:
\begin{equation}
\label{eq: l2_problem_vec}
    \max_{\ba} \sum \ba_i^2 \quad \textrm{s.t. } \: \| \ba - \bb\|_2 \leq \epsilon
\end{equation}

Note that: 
\begin{gather*}
    \| \ba\|^2 = \| \ba - \bb + \bb\|^2 = \| \ba - \bb \|_2^2 + \| \bb\|^2 + 2 \bb^T (\ba - \bb) \\ \leq \epsilon^2 + \| \bb\|_2^2 + 2 \|\bb\|_2 \|\ba - \bb\| \leq \epsilon^2  + \| \bb\|_2^2 + 2\epsilon \| \bb\|_2 = (\|\bb\|_2 + \epsilon)^2
\end{gather*}

Setting $\ba = \bb + \frac{\epsilon\bb}{\|b\|_2}$, we can achieve to the maximum value $\|a\|_2^2 = (\|\bb\|_2 + \epsilon)^2$ without violating the constraints in~\eqref{eq: l2_problem_vec}. Therefore, the optimal solution to Problem~\eqref{eq: l2_problem_eq} is $\Sigma = (\|\Sigma_0\|_2 + \epsilon)^2$ which is $\Tr(\Sigma_0^T\Sigma_0^T) + 2\epsilon\sqrt{\Tr(\Sigma_0^T\Sigma_0^T)} + \epsilon^2$. Therefore, Problem~\eqref{eq: l2_problem_eq} is equivalent to:
\begin{equation}
    \Tr(Q_{\btheta}^T Q_{\btheta}) + 2\epsilon\sqrt{\Tr(Q_{\btheta}^T Q_{\btheta})} + \epsilon^2
\end{equation}
which gives us the desired result in Theorem~\ref{thm: robust_L} part \textbf{b}.

\noindent \textbf{c)} In the case of $p = \infty$, we seek to solve:
\begin{equation}
\label{eq: l_infty_problem}
    \max_{Q} \Tr(Q^T Q) \quad \textrm{s.t. } \: \| \sigma(Q) - \sigma(Q_{\btheta}) \|_{\infty} \leq \epsilon,
\end{equation}
where the uncertainty set is defined on the singular values of the matrix $Q_{\btheta}$.
We can add $\epsilon$ to each singular value of the matrix $Q_{\btheta}$ independently to maximize the above problem. In that case, the optimal solution is $\Tr(Q_{\btheta}^T Q_{\btheta}) + \epsilon^2 + 2\epsilon \Tr(|\Sigma_{\btheta}|)$ , which is the desired result presented in Theorem~\ref{thm: robust_L}. 

\section{Proof of Theorem~\ref{thm: fairness_generalization}}
\label{appendix: generalization}
First, note that there exists $\tilde{\btheta}$ such that ${\rm Tr}(Q_{\tilde{\btheta}}^T Q_{\tilde{\btheta}}) = 0$. To see why, we choose $\tilde{\btheta}$ such that $\hat{y}_{\tilde{\btheta}}(\bx)$ outputs uniformly at random between zero and one independent of $\bx$. Thus, $\hat{y}_{\tilde{\btheta}}(\bx)$ is independent of the sensitive attribute $s$. By the definition of the Exponential \Renyi Mutual Information (ERMI), ${\rm Tr}(Q_{\tilde{\btheta}}^T Q_{\tilde{\btheta}}) = 0$, since $\hat{y}_{\tilde{\btheta}}(\bx)$ and $s$ are independent. Assume that $\btheta^{*}_{\lambda, \epsilon}$ is the optimal solution of Problem~\ref{eq: fermi_fair_empirical_risk} for the given $\lambda$ and $\epsilon$. Since, the  $\btheta^*_{\lambda, \epsilon}$ is the optimizer of~\eqref{eq: fermi_fair_empirical_risk}, we have:
\begin{equation}
        \mathbb{E}_{\mathbb{P}_{\textrm{tr}}}[\ell(\hat{y}_{\btheta^{*}_{\lambda, \epsilon}}(\bx), y)] + \lambda {\rm Tr}(Q_{\btheta^{*}_{\lambda, \epsilon}}^T Q_{\btheta^{*}_{\lambda, \epsilon}})  \leq   \mathbb{E}_{\mathbb{P}_{\textrm{tr}}}[\ell(\hat{y}_{\tilde{\btheta}}(\bx), y)]  + \lambda {\rm Tr}(Q_{\tilde{\btheta}}^T Q_{\tilde{\btheta}}) = \mathbb{E}_{\mathbb{P}_{\textrm{tr}}}[\ell(\hat{y}_{\tilde{\btheta}}(\bx), y)]  
\end{equation}
The equality holds, since the ${\rm Tr}(Q_{\tilde{\btheta}}^T Q_{\tilde{\btheta}})$ is zero. Therefore:
\begin{equation}
        {\rm Tr}(Q_{\btheta^{*}_{\lambda, \epsilon}}^T Q_{\btheta^{*}_{\lambda, \epsilon}})  \leq   \Big(\mathbb{E}_{\mathbb{P}_{\textrm{tr}}}[\ell(\hat{y}_{\tilde{\btheta}}(\bx), y)] -  \mathbb{E}_{\mathbb{P}_{tr}}[\ell(\hat{y}_{\btheta^{*}}(\bx), y)]\Big) / \lambda   
\end{equation}
Since the cross-entropy loss is always greater than or equal to zero, we have:
\begin{equation*}
    \mathbb{E}_{\mathbb{P}_{\textrm{tr}}}[\ell(\hat{y}_{\btheta^{*}}(\bx), y)] \geq 0
\end{equation*}
Further, assume that the loss value at point $\tilde{\btheta} = B$. Thus:
\begin{equation}
    {\rm Tr}(Q_{\btheta^{*}_{\lambda, \epsilon}}^T Q_{\btheta^{*}_{\lambda, \epsilon}})  \leq \frac{\mathbb{E}_{\mathbb{P}_{\textrm{tr}}}[\ell(\hat{y}_{\tilde{\btheta}}(\bx), y)]}{\lambda} \leq \frac{B}{\lambda}   
\end{equation}
Therefore, if we choose $\lambda \geq \frac{B}{\gamma}$, we have
\begin{equation}
        {\rm Tr}(Q_{\btheta^{*}_{\lambda, \epsilon}}^T Q_{\btheta^{*}_{\lambda, \epsilon}}) \leq \gamma  
\end{equation}
We showed the above inequality for arbitrary choice of $\epsilon$.  Next, we choose $\epsilon$ such that $Q_{\btheta^{*}_{\epsilon, \lambda}}^{\textrm{test}} \in \mathcal{B}(Q_{\btheta^{*}_{\epsilon, \lambda}}^{\textrm{train}}, \epsilon)$. Note that according to \citet{witsenhausen1975sequences}, the largest singular value of the matrix $Q$ is one. As a result, the size of all eigenvalues of matrices $Q_{\btheta^{*}_{\epsilon, \lambda}}^{\textrm{test}}$ and $Q_{\btheta^{*}_{\epsilon, \lambda}}^{\textrm{train}}$ are less than or equal to $1$. Therefore, by choice of large enough $\epsilon$, the $L_p$ distance of the vectors of eigenvalues for $Q_{\btheta^{*}_{\epsilon, \lambda}}^{\textrm{test}}$ and $Q_{\btheta^{*}_{\epsilon, \lambda}}^{\textrm{train}}$ is within $\epsilon$.  
Hence:
\begin{equation*}
{\rm Tr}(Q_{\btheta}^{(\textrm{test}) T} Q_{\btheta}^{\textrm{test}}) \leq \max_{Q \in \mathcal{B}(Q_{\btheta}^{\textrm{train}}, \epsilon)} {\rm Tr}(Q^T Q) < \gamma.   
\end{equation*}
According to~\citet[Lemma 3]{lowy2021fermi}, the exponential \Renyi Mutual Information is an upper bound for $L_{\infty}$ fairness violation (demographic parity violation). Thus:
\begin{equation*}
    \textrm{DPV} = | P(\hat{y} = 1 | s = 1) - P(\hat{y} = 1 | s = 0) | < \gamma,
\end{equation*}
which completes the proof.

\section{Proof of Lemma~\ref{lemma:stochastic_reformulation}}
\label{appendix: B}
First, we show that for any $z > 0$:
\begin{equation}
\label{eq:sqrt_z}
    \sqrt{z} = \min_{\alpha > 0 } \:  \frac{1}{2}(z\alpha + \frac{1}{\alpha})
\end{equation}
Let $g_z(\alpha) = \frac{1}{2}(z\alpha + \frac{1}{\alpha})$. Since $\frac{\partial^2 g_z(\alpha)}{\partial \alpha^2} =  \frac{1}{\alpha^3} > 0$, the function $g_z$ is strictly convex with respect to $\alpha$. Thus, the minimizer is unique. The minimizer of $g_z(\alpha)$ can be obtained by setting the derivative to zero with respect to $\alpha$:
\begin{gather*}
 \frac{\partial g_z(\alpha)}{\partial \alpha} = z + \frac{1}{\alpha^2} = 0 \rightarrow 
 \alpha^* = \frac{1}{\sqrt{z}}\end{gather*}
 Therefore, the optimal value can be obtained by plugging in $\alpha^*$ into $g_z$:
 \begin{gather*}
 g_z(\alpha^*) = \frac{1}{2}(\frac{z}{\sqrt{z}} + \frac{1}{\frac{1}{\sqrt{z}}}) = \frac{1}{2}(\sqrt{z} + \sqrt{z}) = \sqrt{z},
\end{gather*}
that proves~\eqref{eq:sqrt_z}.
Based on~\citep[Lemma 1]{lowy2021fermi}, 
\begin{equation}
    {\rm Tr}(Q^T Q) = \Psi(\btheta, W) = \frac{1}{N} \sum_{i=1}^{N} \psi(\bz_i; \btheta, W)
\end{equation}
where $\psi$ is given by Equation~\eqref{eq: psi}. Thus Problem~\eqref{eq:Robust_FERMI_L2} can be written as:
\begin{equation}
\label{eq:N_samples}
\frac{1}{N}\sum_{i=1}^N \ell(\hat{y}_{\btheta}(\bx_i), y_i) +\lambda \Psi(\btheta, W) + 2\lambda \epsilon \sqrt{\Psi(\btheta, W)}
\end{equation}
Applying Equation~\eqref{eq:sqrt_z} to $\sqrt{\Psi(\btheta, W)}$, we have: 
\begin{equation}
\label{eq:sqrt_psi}
\sqrt{\Psi(\btheta, W)} = \frac{1}{2}\Big(\min_{\alpha>0} \alpha \Psi(\btheta, W) + \frac{1}{\alpha}\Big) 
\end{equation}
Combining~\eqref{eq:N_samples} and~\eqref{eq:sqrt_psi}, Problem~\eqref{eq:Robust_FERMI_L2} can be reformulated as:
\begin{align}
\label{eq:final_formulation}
\min_{\boldsymbol{\theta}} \: &\max_{W \in \mathcal{W}} \frac{1}{N}\sum_{i=1}^N \ell(\hat{y}_{\btheta}(\bx_i), y_i) +\lambda \Psi(\btheta, W) \nonumber \\ &+ \min_{\alpha > 0} \: \lambda \epsilon\alpha \Psi(\btheta, W) + \frac{\lambda \epsilon}{\alpha}
\end{align}
Note that the maximization problem with respect to $W$ is concave due to the concavity of $\Psi$, and the minimization problem with respect to $\alpha$ is convex on $\alpha > 0$. To switch the minimum and maximum terms, we need to have boundedness on the parameters of the minimization problem. Obviously, $\alpha$ is bounded below by $0$. Further, note that the optimal $\alpha$ satisfies the following condition due to the first-order stationarity condition: 
\begin{gather*}
\Psi(\btheta, W) - \frac{1}{\alpha^{*2}} = 0 \rightarrow \alpha^{*} = \frac{1}{\sqrt{\Psi(\btheta, W)}}
\end{gather*}
Since $\Psi(\btheta, W)$ is the reformulation of $\Tr(Q_{\btheta}^TQ_{\btheta})$ and the latter term is the summation of all singular values, it is always greater than or equal to $1$ (since the largest singular value equals $1$). As a result $\alpha^* \leq 1$. Therefore, it is bounded from above as well. Thus,  we can switch the maximization and minimization problems in~\eqref{eq:final_formulation} due to the minimax theorem~\citep{sion1958general}. Hence, Problem~\eqref{eq:Robust_FERMI_L2} is equivalent to 
\begin{align*}
\min_{\boldsymbol{\theta}, \alpha > 0} \: \max_{W} \frac{1}{N}\sum_{i=1}^N \ell(\hat{y}_{\btheta}(\bx_i), y_i) +\lambda \Psi(\btheta, W) + \lambda \epsilon \alpha \Psi(\btheta, W) + \frac{\lambda\epsilon}{\alpha},
\end{align*}
which is precisely Problem~\eqref{eq:robust_l2_reformulation}.

\section{Significance and Limitations of Theorem~\ref{thm: fairness_generalization}}
\label{appendix: generalization_limitations}
The obtained result is based on the two crucial properties of~\eqref{eq: dro_fair_risk_minimization} formulation: First, bounding the ERMI between sensitive attributes and predictions results in bounding the demographic parity violation. Second, the uncertainty sets characterized by $L_p$ balls around the matrix $Q^{\textrm{train}}$ can be chosen such that the current distribution shift lies within the uncertainty ball. In that sense, the proposed framework is less restrictive than the aforementioned related works that only focus on a specific type of shift, such as demographic, covariate, or label shifts. Note that Theorem~\ref{thm: fairness_generalization} only provides a generalization guarantee on the unseen data for the model's fairness. Therefore, it does not guarantee any level of accuracy on the test data.  An open question is whether optimizing the loss function in~\eqref{eq:Robust_FERMI_L2} leads to a solution with both Fairness and Accuracy guarantees on the unseen data. Such a problem is challenging even in the case of generalizability of non-convex trained models when no fairness concerns are considered. Further, the satisfaction of $Q^{\textrm{test}}_{\btheta} \in \mathcal{B}(Q^{\textrm{tr}}_{\btheta})$ at each iteration as a necessary condition depends on the choice of $\epsilon$. A large distance between $Q^{\textrm{test}}_{\btheta}$ and $Q^{\textrm{tr}}_{\btheta}$ can lead to larger $\epsilon$, and therefore the poor performance of the trained model in terms of accuracy as the compensation for promoting fairness.

\vspace{-3mm}
\section{Derivation of Algorithm~\ref{alg: robust_l1} and Details on Computing Gradients}
Algorithm~\ref{alg: robust_l1} requires to update the probability matrices $\hat{P}_{\hat{y}}$ and $\hat{P}_{\hat{y}, s}$ after updating $\btheta$. Assume that the output of the classification model is a probability vector (logits in neural networks or logistic regression) where the probability of assigning label $j$ to the data point $\bx$ is $F_{j}(\bx; \btheta)$. Note that $\hat{y}_{\btheta}(\bx) = \argmax_{j} F_{j}(\bx; \btheta)$. One can compute the elements of probability matrices as follows:
\begin{gather}
    \hat{P}_{\hat{y}} [j][j] = \mathbb{P}(\hat{y}_{\btheta} = j) = \frac{1}{n} \sum_{i=1}^{n} F_{j}(\bx; \btheta) \label{eq: grad1}\\ 
    \hat{P}_{\hat{y}, s}[j][k] = \mathbb{P}(\hat{y}_{\btheta} = j, s=k) = \mathbb{P}(s=k) \mathbb{P}(\hat{y}_{\btheta} = j | s=k) =  \frac{\pi_k}{n} \sum_{i=1}^n F_{j}(\bx_i; \btheta) \mathbbm{1}(s_i = k) \label{eq: grad2}
\end{gather}
As a note, $\mathbb{P}(s = k) = \pi_k = \frac{1}{n} \sum_{i=1}^n \mathbbm{1}(s_i = k)$ and it is constant through the algorithm.

Next, we show how to compute the gradient of $\Tr(Q_{\btheta}^T Q_{\btheta})$ with respect to $\btheta$. First, it is evident that:
\begin{equation*}
    \Tr(Q_{\btheta}^T Q_{\btheta}) = \sum_{j} \sum_{k} \frac{\mathbb{P}^2(\hat{y}_{\btheta} = j, s = k)}{\mathbb{P}(\hat{y}_{\btheta} = j) \mathbb{P}(s = k)}
\end{equation*}
Thus, its gradient with respect to $\btheta$ using the chain rule equals:
\begin{equation*}
    \sum_{j, k} \frac{2\mathbb{P}(\hat{y}_{\btheta} = j, s = k) \mathbb{P}(\hat{y}_{\btheta} = j) \nabla_{\btheta} \mathbb{P}(\hat{y}_{\btheta} = j, s = k) - \mathbb{P}^2(\hat{y}_{\btheta} = j, s = k) \nabla_{\btheta} \mathbb{P}(\hat{y}_{\btheta} = j)}{\mathbb{P}^2(\hat{y}_{\btheta} = j) \mathbb{P}(s = k)}
\end{equation*}
where the gradient of $\mathbb{P}(\hat{y}_{\btheta} = j, s = k)$ can be computed using Equations~\eqref{eq: grad1} and~\eqref{eq: grad2}:
\begin{gather}
\nabla_{\btheta} \mathbb{P}(\hat{y}_{\btheta} = j) = \frac{1}{n} \sum_{i=1}^{n} \nabla_{\btheta} F_{j}(\bx; \btheta) \\
    \nabla_{\btheta} \mathbb{P}(\hat{y}_{\btheta} = j, s = k) = \frac{\pi_k}{n} \sum_{i=1}^n \nabla_{\btheta} F_{j}(\bx_i; \btheta) \mathbbm{1}(s_i = k)
\end{gather}
\label{appendix: gradient_computation}
\vspace{-1mm}

\vspace{-3mm}
\section{Stochastic Gradient Descent Ascent (SGDA) Algorithms 
\label{appendix: stochastic}
For~\eqref{eq: fermi_fair_empirical_risk}}
This section proposes the stochastic algorithm for solving~\eqref{eq: fermi_fair_empirical_risk} under the $L_2$ norm ball as the uncertainty set. Then, we show the algorithm finds a stationary solution to the problem in $\mathcal{O}(\frac{1}{\epsilon^5})$. As mentioned in Section~\ref{sec: stochastic},~\eqref{eq: fermi_fair_empirical_risk} can be rewritten as:
\begin{equation*}
    \min_{\alpha>0, \boldsymbol{\theta}} \: \max_{W \in \mathcal{W}} \frac{1}{n} \big[ \sum_{i=1}^{n} \ell(\bz_i; \btheta) + \lambda(1 + \epsilon \alpha) \psi(\bz_i; \btheta, W) \big] + \frac{\lambda\epsilon}{\alpha}
\end{equation*}
Since the objective function is represented as an average/expectation over $n$ data points, the gradient vector with respect to $\btheta$ evaluated for a given batch of data is an unbiased estimator of the gradient with respect to all data points. Further, the problem is strongly concave with respect to the matrix $W$. Therefore, a standard stochastic gradient descent ascent (SGDA) algorithm will find a stationary solution to the problem. Algorithm~\ref{alg: stochastic_robust_l2} describes the procedure of optimizing~\eqref{eq:robust_l2_reformulation}.
\vspace{-1mm}
\begin{algorithm}
    \caption{Stochastic Distributionally Robust FERMI Under $L_2$ Ball Uncertainty}
    \label{alg: stochastic_robust_l2}
    \begin{algorithmic}[1]
	 \STATE \textbf{Input}: $\btheta^0, W^0,\alpha^0$, step-sizes $\eta$, fairness parameter $\lambda \geq 0,$ number of iterations $T$, uncertainty ball radius $\epsilon$. 
  \vspace{2mm}
    \FOR {$t = 1, \ldots, T$}
    \vspace{2mm}
    \STATE Take a mini-batch of data $\mathcal{B}_t\subseteq \{1,\ldots, N\}$ uniformly at random
    \vspace{2mm}
    \STATE $\small \btheta^{t} = \btheta^{t-1} - \eta \Big[ \sum_{\bz \in \mathcal{B}_t} \nabla_{\btheta} \ell(\bz; \btheta^{t-1}) + \sum_{\bz \in \mathcal{B}_t} \nabla_{\btheta} \psi(\bz; \btheta^{t-1}, W^{t-1})\Big]$
    \vspace{2mm}
    \STATE $\alpha^t = \alpha^{t-1} - \eta\lambda \epsilon \Big[\sum_{\bz \in \mathcal{B}_t} \psi(\bz; \btheta^{t-1}, W^{t-1}) -  1/\alpha_{t-1}^2\Big]$ 
    \vspace{2mm}
    \STATE $W^{t} = \Pi_{\mathcal{W}}\Big[W^{t-1} + \eta\lambda(1+\epsilon\alpha^{t-1}) \sum_{\bz \in \mathcal{B}_t} \nabla_{W} \psi(\bz; \btheta^{t-1}, W^{t-1})\Big]$
    \ENDFOR
    \STATE \textbf{Return:} $\btheta^T$. 
\end{algorithmic}
\vspace{-.03in}
\end{algorithm}

Next, we prove the convergence of Algorithm~\ref{alg: stochastic_robust_l2} to a stationary solution of Problem~\eqref{eq:Robust_FERMI_L2}. 
\begin{proposition}
\label{thm: stochastic_convergence}
    Assume that $\ell(\cdot, \cdot)$ and $\mathcal{F}(\bx, \btheta)$, the probabilistic output of the model (logits) are Lipschitz continuous and differentiable with the Lipschitz gradient. Further, assume that $P_{s}(s = i) > 0 \quad \forall i \in \mathcal{S}$, and $P_{\hat{y}}(\hat{y}_{\btheta} = j) > 0 \quad \forall \btheta$ and $\forall j\in \mathcal{Y}$. Then, for any batch size of $1 \leq b \leq n$, Algorithm~\ref{alg: stochastic_robust_l2} finds an $\epsilon$-first order stationary point of Problem~\eqref{eq:Robust_FERMI_L2} in $T = \mathcal{O}(\frac{1}{\epsilon^5})$ iterations. 
\end{proposition}
The proof is based on the following assumptions:
\begin{itemize}
    \item $P_s(s = i) > 0 \quad \forall i \in \mathcal{S}$ 
    \item $P_{\hat{y}} (\hat{y} = j) > 0 \quad \forall j \in \mathcal{Y}$ for every iteration of the algorithm. 
    \item The loss function $\ell(\bx, \btheta)$ and the probabilistic output $\mathcal{F}(\bx, \btheta)$ are Lipschitz continuous with the Lipschitz constant $L$. 
    \item The loss function $\ell(\bx, \btheta)$ and the probabilistic output $\mathcal{F}(\bx, \btheta)$ are $\beta$-smooth, meaning that their gradients are Lipschitz continuous with the Lipschitz constant $\beta$.
\end{itemize}
\begin{remark}
Note that the first two assumptions are true in practice. Because we assume at least one sample from each sensitive group should be available in the training data. Further, the probability of predicting any label in every iteration should not be zero. In our experiments, such an assumption always holds. If, for some extreme cases, one probability goes to zero, one can add a small perturbation to $\btheta$ to avoid the exact zero probability. The third and fourth assumptions are standard in the convergence of iterative method proofs, and they hold for loss functions such as cross-entropy and mean-squared loss over bounded inputs.      
\end{remark}
\begin{proof}
Define $\btheta^{'} = [\btheta, \alpha]$, which means $\btheta^{'}$ is obtained by adding the parameter $\alpha$ to vector $\btheta$. In that case, Algorithm~\ref{alg: stochastic_robust_l2} can be seen as one step of gradient descent to $\btheta^{'}$ and one step of projected gradient ascent to $W$. Note that, although the maximization problem is originally unconstrained, we are required to apply the projection to the set $\mathcal{W} = \{W \in \mathbb{R}^{m\times k}: \|W\|_F \leq \frac{2}{\hat{P}_{\hat{y}}^{\min} \sqrt{\hat{P}_{s}^{\min}}}\}$ to ensure the Lipschitz continuity and the boundedness of the variance of the gradient estimator. The convergence of Algorithm~\ref{alg: stochastic_robust_l2} to an $\epsilon$-stationary solution of Problem~\ref{eq:Robust_FERMI_L2} in $\mathcal{O}(\frac{1}{\epsilon^5})$ is the direct result of Theorem 3 in~\citet{lowy2021fermi}.    
\end{proof}

\vspace{-3mm}
\section{Convergence of Algorithm~\ref{alg: robust_l1}}
\vspace{-1mm}
\label{appendix:ConvergenceofDeterministic}
Since we solve the maximization problem in~\eqref{eq:robust_l1}
in closed form (see lines 5 and 6 in Algorithm~\ref{alg: robust_l1}), as an immediate result of Theorem 27 in~\citet{jin2020local}, Algorithm~\ref{alg: robust_l1} converges to a stationary solution of Problem~\eqref{eq:robust_l1} in $\mathcal{O}(\frac{1}{\epsilon^4})$. Note that, although the problem is non-convex non-concave, we have a max oracle (we can exactly compute the solution to the maximum problem in closed form). 

\begin{theorem}\label{thm: Fair-classification-}
\label{thm: convergence}
Suppose that the objective function defined in~\eqref{eq: fermi_fair_empirical_risk} is $L_0$-Lipschitz and $L_1$-gradient Lipschitz for the choice of $p=1$ in Theorem~\ref{thm: robust_L}. Then, Algorithm~\ref{alg: robust_l1} computes an $\epsilon$-stationary solution of the objective function in \eqref{eq:robust_l1} in  $\mathcal{O} (\epsilon^{-4})$ iterations.
\end{theorem}

\vspace{-2mm}
\section{Full-Batch Algorithms for Problem~\eqref{eq:Robust_FERMI_L2} and Problem~\eqref{eq:Robust_FERMI_L_inf}}
\vspace{-1mm}
\label{app: D}
To solve Problem~\ref{eq:Robust_FERMI_L2}, we use the following gradient descent-ascent (GDA) approach:

\vspace{-1mm}
\begin{algorithm}
    \caption{Distributionally Robust FERMI under $L_2$ Ball Uncertainty}
    \label{alg: robust_l2}
    \begin{algorithmic}[1]
	 \STATE \textbf{Input}: $\btheta^0 \in \mathbb{R}^{d_{\theta}}, ~W^0 = 0$, step-sizes $\eta$, fairness parameter $\lambda \geq 0,$ iteration number $T$.
    
    \FOR {$t = 1, \ldots, T$}
    
    \STATE $\btheta^{t} = \btheta^{t-1} - \eta \nabla_{\btheta} f_2 (\btheta^{t-1})$ 
    \vspace{1mm}
        
    \STATE $\small \rho_W = -\Tr \Big(W^{t-1} \widehat{P}_{\hat{y}} W^{t-1, T} +2W^{t-1} \widehat{P}_{\hat{y}, s}
    \widehat{P}_{s}^{-1/2} \Big) $
    \vspace{1mm}
    \STATE Set $R_W = \lambda \rho_W + 2\lambda\epsilon \sqrt{\rho_W}$
    \vspace{1mm}
    \STATE Set $W^t = W^{t-1} + \eta \nabla_{W} R_W$ 
    
    \vspace{1mm}
    \STATE Update $\widehat{P}_{\hat{y}}$ and $\widehat{P}_{\hat{y}, s}$ 
    
    \ENDFOR
    \STATE \textbf{Return:} $\btheta^T, W^T.$
\end{algorithmic}
\vspace{-.03in}
\end{algorithm}

\begin{remark}
The convergence of Algorithm~\ref{alg: robust_l2} is obtained by setting the batch size in Theorem~\ref{thm: stochastic_convergence} to $|B| = N$.    
\end{remark}

To solve Problem~\eqref{eq:Robust_FERMI_L_inf}, we use the following algorithm:
\begin{algorithm}
    \caption{Distributionally Robust FERMI under $L_{\infty}$ Ball Uncertainty}
    \label{alg: robust_l_infty}
    \begin{algorithmic}[1]
	 \STATE \textbf{Input}: $\btheta^0 \in \mathbb{R}^{d_{\theta}}, ~W^0 = 0$, step-sizes $\eta$, fairness parameter $\lambda \geq 0,$ iteration number $T$.
    
    \FOR {$t = 1, \ldots, T$}
    
       \STATE \small $\btheta^{t} = \btheta^{t-1} - \eta \Big(\frac{1}{n}\sum_{i=1}^n \nabla_{\btheta} \ell(\hat{y}_{\btheta^{t-1}}(\bx_i), y) + 2\lambda \epsilon \Tr(|\Sigma_{\btheta^{t-1}}|) + \lambda \Tr \big(\nabla_{\btheta}(Q_{\btheta^{t-1}}^T Q_{\btheta^{t-1}})\big)\Big)$ 
    \vspace{1mm}
    \ENDFOR
    \STATE \textbf{Return:} $\btheta^T$
\end{algorithmic}
\vspace{-.03in}
\end{algorithm}

\section{Algorithms for CVaR-Dr. CVaR and GROUP-Dr. FERMI}
\label{appendix: cvar_alg}
By a simple modification of Algorithm~\ref{alg: robust_l1}, one can derive convergent algorithms to the stationary points of~\eqref{eq: dro_cvar} and~\eqref{eq: dro_group}. Note that the maximization step at each iteration remains the same. The only change is in the minimization part. Instead of minimizing with respect to $\btheta$, we need to update the extra parameter $\eta$ as an optimization problem using (sub-)gradient descent. Algorithm~\ref{alg: cvar} describes the procedure of optimizing~\eqref{eq: dro_cvar}. At each iteration, we apply one step of (sub-)gradient descent to $\btheta$ and $\eta$ and then update $\bv$ and $W$ in closed form.
\begin{algorithm}
    \caption{CVaR-Dr. FERMI Algorithm}
    \label{alg: cvar}
    \begin{algorithmic}[1]
	 \STATE \textbf{Input}: $\btheta^0 \in \mathbb{R}^{d_{\theta}}, ~W^0 = 0$, step-sizes $\alpha$, fairness parameter $\lambda \geq 0,$ iteration number $T$.
    
    \FOR {$t = 1, \ldots, T$}
    
    \STATE \small $\btheta^{t} = \btheta^{t-1} - \frac{\alpha}{n}\sum_{i=1}^n \partial_{\btheta} [\ell(\hat{y}_{\btheta^{t-1}}(\bx_i), y) - \eta_t]_{+} + \frac{\lambda \epsilon}{\alpha}  \partial_{\btheta}\sqrt{\bv^T\big(Q_{\btheta^{t-1}}^T Q_{\btheta^{t-1}}\big)\bv} + \frac{\lambda}{\alpha} \Tr \big(\partial_{\btheta}(Q_{\btheta^{t-1}}^T Q_{\btheta^{t-1}}\big)\Big) $ 
    \vspace{1mm}

    \STATE \small $\eta^{t} = \eta^{t-1} - \frac{\alpha}{n} \sum_{i=1}^n \Big( \partial_{\eta}[\ell(\hat{y}_{\btheta^{t-1}}(\bx_i), y) - \eta_t]_{+} \Big) - \alpha$ 
    \vspace{1mm}
    
    \STATE Set \small $W^{t} = \widehat{P}_{s}^{-1/2} \widehat{P}_{\hat{y}, s}^T \widehat{P}_{\hat{y}}^{-1}$
    \vspace{1mm}

    \STATE Set $\bv$ to the second largest singular vector of $Q_{\btheta^t}$ by performing SVD on $Q_{\btheta^t}$.
    \vspace{1mm}
    
    \STATE Update $\widehat{P}_{\hat{y}}$ and $\widehat{P}_{\hat{y}, s}$ as a function of $\btheta^{t-1}$. 
    \ENDFOR
    \STATE \textbf{Return:} $\btheta.$
\end{algorithmic}
\vspace{-.03in}
\end{algorithm}
Similarly, one can optimize~\eqref{eq: dro_group} by applying gradient descent to the $\btheta$ as in Algorithm~\ref{alg: robust_l1} and the group DRO parameters as in Algorithm 1 in~\citet{sagawa2019distributionally} and then applying $\bv$ and $W$ in the closed form at each iteration. 

\section{Performance of Stochastic DR ERMI}
\label{app: stochastic_plot}
In this section, we evaluate Algorithm~\ref{alg: stochastic_robust_l2} for different batch sizes. To this end, we learn fair models with different batch sizes on the adult dataset where the minority group is under-represented. We compare Algorithm~\ref{alg: stochastic_robust_l2} to~\citep{lowy2021fermi},~\citep{baharlouei2019renyi},~\citep{rezaei2021robust},~\citep{mary2019fairness}, and~\citep{cho2020fair} as the baselines supporting stochastic updates. Since the gradient estimator in Algorithm~\ref{alg: stochastic_robust_l2} is unbiased, the performance of Algorithm~\ref{alg: stochastic_robust_l2} remains consistently well for different batch sizes. However, by reducing the batch size, other methods (except~\citet{lowy2021fermi}) fail to keep their initial performance on the full batch setting. Comparing our approach and~\citet{lowy2021fermi}, we achieve better generalization in terms of the fairness-accuracy tradeoff for different batch sizes.

\begin{figure}[H]
\vspace{-12mm}
    \begin{center}    \centerline{\includegraphics[width=1\columnwidth]{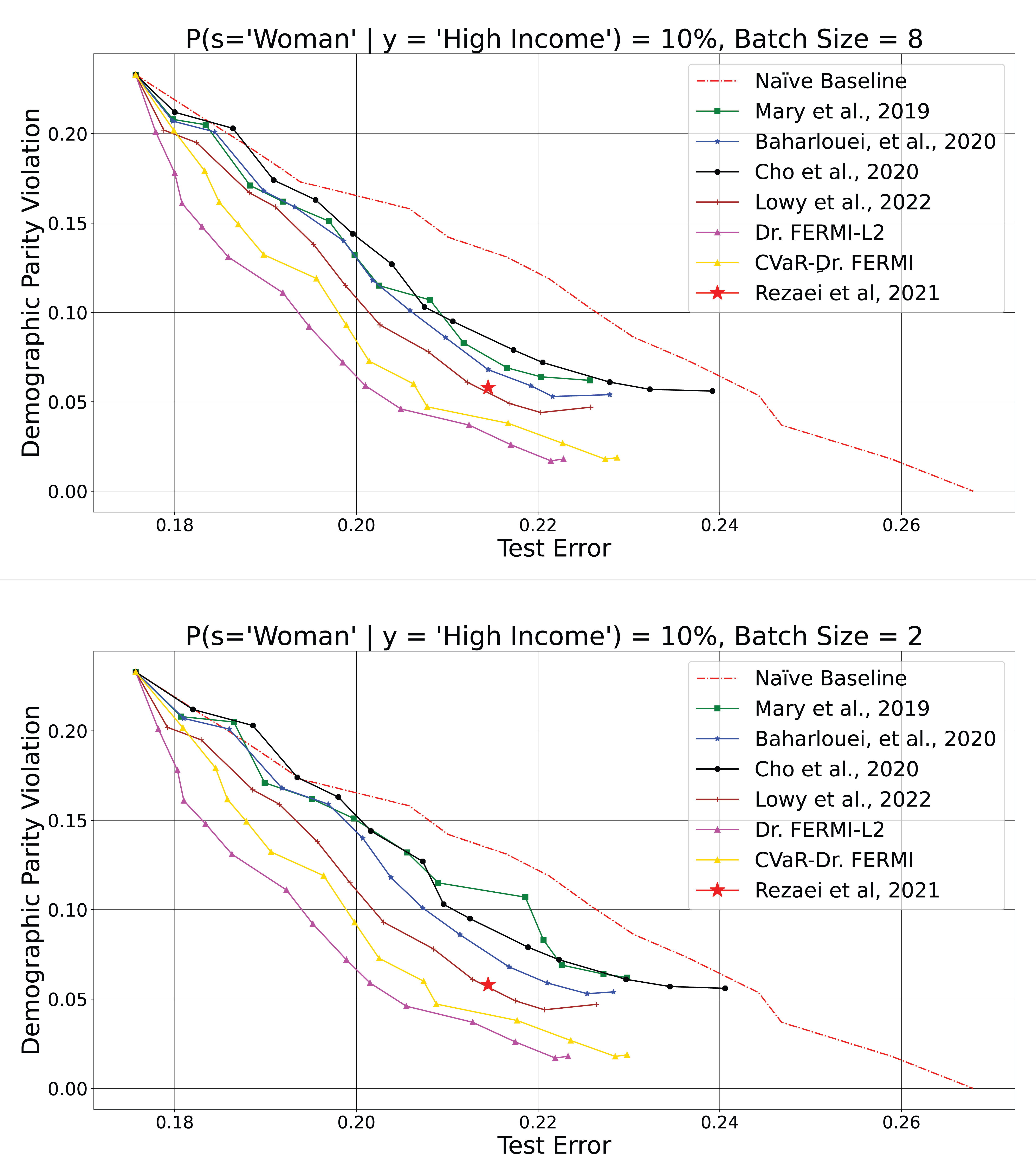}}
    \caption{\small Performance of different methods regarding fairness and accuracy for small batch sizes of $2$ and $8$. Our robust formulation works consistently well even for small batch sizes due to the unbiasedness of the gradient, while other methods are more sensitive to smaller batch sizes.}
    \label{fig: stochastic}
\end{center}
\end{figure}

\newpage
\section{More Details on Experiments}
\label{app:AdditionalExperiments}
This section presents a more detailed version of Table~\ref{tab: non_binary} with added accuracy and fairness standard deviations among $50$ states.

\begin{table*}[h]
\label{tab: non_binary_std}
\vspace{-3mm}
\centering
\begin{tabular}{|c|c|c|c|}
\hline
\textbf{Method}         & \textbf{Tr Accuracy} & \textbf{Test Accuracy} & \textbf{Test EO Violation} \\ \hline
Mary et al., 2019 & $71.41 \pm 1.61\%$  & $68.35 \pm 1.69\%$ & $0.1132 \pm 0.023$  \\ \hline
Cho et al., 2020 & $71.84 \pm 1.55\%$ & $68.91 \pm 1.57\%$ & $0.1347 \pm 0.027$ \\ \hline
Baharlouei et al., 2020 & $72.77 \pm 1.37\%$ & $69.44 \pm 1.39\%$ & $0.0652 \pm 0.013$                  \\ \hline
Lowy et al., 2022 & $73.81 \pm 0.98\%$ & $70.22 \pm 1.01\%$ & $0.0512 \pm 0.009$ \\ \hline
\textbf{Dr. FERMI-$L_1$} & $73.45 \pm 1.02\%$  & $70.09 \pm 1.06\%$ & $0.0392 \pm 0.004$ \\ \hline
\textbf{Dr. FERMI-$L_2$} & $73.12 \pm 1.04\%$ & $69.71 \pm 1.11\%$ & $\textbf{0.0346} \pm 0.005$         \\ \hline
\textbf{Dr. FERMI-$L_{\infty}$} & $73.57 \pm 0.96\% $ & $69.88 \pm 0.99\%$ & {\ul 0.0359}$\pm 0.008$                     \\ \hline
\textbf{CVaR-Dr. FERMI-$L_1$} & $\textbf{74.21} \pm 0.24\%$  &  $\textbf{70.94} \pm 0.28\%$ & $0.0471 \pm 0.011$ \\ \hline
\textbf{CVaR-Dr. FERMI-$L_2$} & $73.84 \pm 0.28\%$ & $70.26 \pm 0.31\%$ & $0.0429 \pm 0.008$ \\ \hline
\textbf{CVaR-Dr. FERMI-$L_{\infty}$} & {\ul 73.92}$\pm 0.31\%$ & {\ul 70.45}$\pm 0.37\%$  & $0.0466 \pm 0.013$ \\ \hline
\end{tabular}
\vspace{-1mm}
\caption{\small Train and Test Accuracy and Fairness Violation of Different Methods on ACS PUBS dataset. The reported numbers are the average and the standard deviation of the values among $50$ states.}
\end{table*}
Further, to show the memory efficiency of Dr. FERMI, we report the memory consumption, the execution time
\begin{table}[H]
\centering
\begin{tabular}{|c|c|c|}
\hline
Method                   & Memory Consumption      & Training Time \\ \hline
CVaR Dr. FERMI            & {\ul \textless{}800 Mb} & 783 (s)       \\ \hline
Dr. FERMI                 & \textless{}800 Mb       & 692 (s)       \\ \hline
Lowy et al, 2022         & \textless{}800 Mb       & 641 (s)       \\ \hline
Cho et al., 2021         & 1.21 Gb                 & 3641 (s)      \\ \hline
Rezaei et al., 2021      & 1.59 Gb                 & 3095 (s)      \\ \hline
Baharlouei, et al., 2020 & \textless 900 Mb        & 2150 (s)      \\ \hline
Mary, et al., 2019 & \textless 1.43 Gb        & 1702 (s)      \\ \hline
\end{tabular}
\vspace{3mm}
\caption{ Memory Consumption and Computation Time of Different Approaches in Figure 1. Note that the reported values are for the batch size of $8$. The required memory for the full batch algorithm is more than $4$GB, and the required time is nearly the same (since the number of epochs is the same). Note that while other approaches lose performance due to the smaller batch sizes, Dr. FERMI and CVar Dr. FERMI preserve their performance even for small batch sizes (as small as $1$).}
\end{table}

\section{Hyper-Parameter Tuning}
\label{app:HyperParamCode}
DP-FERMI has two hyper-parameters $\epsilon$ and $\lambda$. Further, the presented algorithms in the paper have a step-size (learning rate) $\eta$, and the number of iterations $T$. We use $\eta = 10^{-5}$ and $T = 3000$ in the code. These values are obtained by changing $\eta$ and $T$ over $10$ different values in different runs on ACS PUMS~\citep{ding2021retiring} data. We consider two scenarios for tuning $\lambda$  and $\epsilon$. If a sample of data from the target domain is \textbf{available} for the validation, we reserve that data as the validation set to choose the optimal $\lambda \in \{0.1, 0.5, 1, 2, 5, 10, 20, 50\}$ and $\epsilon \in \{0.01, 0.02, 0.05, 0.1, 0.2, 0.5, 1, 2, 5, 10\}$. In the second scenario, when no data from the target domain is provided, one can rely on the $k$-fold cross-validation on the source data. A more resilient approach is to create the validation dataset by oversampling the minority groups. Based on this idea, we do weighted sampling based on the population of sensitive groups (oversampling from minorities), and then, we choose the optimal $\lambda$ and $\epsilon$ as in scenario $1$. 

\section{Limitations}
\label{app:Limitations}
In this paper, we address the fair empirical risk minimization in the presence of the distribution shift by handling the distribution shift of the target domain for the fairness and accuracy parts separately. As shown in Theorem~\ref{thm: fairness_generalization}, we can bound the fairness violation of the target domain with the proper choice of hyper-parameters. However, the proposed framework has several limitations. First, the defined uncertainty set for the fairness term is limited to  $L_p$ norm balls (in certain variables) and the Exponential \Renyi Mutual Information (ERMI) as the measure of fairness. Second, the presented guarantee is just for the fairness violation and does not offer any guarantee for the fairness-accuracy tradeoff. Moreover, the guarantees are for the global optimal solutions, which may not be possible to compute due to the nonconvexity of the problem. 

\section{Broader Impacts}
\label{app:BI}


The emergence of large-scale machine learning models and artificial intelligence tools has brought about a pressing need to ensure the deployment of safe and reliable models. Now, more than ever, it is crucial to address the challenges posed by distribution shifts and strive for fairness in machine learning. This paper presents a comprehensive framework that tackles the deployment of large-scale fair machine learning models, offering provably convergent stochastic algorithms that come with statistical guarantees on the fairness of the models in the target domain.

While it is important to note that our proposed algorithms may not provide an all-encompassing solution for every application, we believe that sharing our research findings with the broader community will positively impact society. By publishing and disseminating our work, we hope to contribute to the ongoing discourse on fairness in machine learning and foster a collaborative environment for further advancements in this field. Moreover, the significance of our algorithms extends beyond the context of fairness. They can potentially be harnessed in a wide range of applications to enforce statistical independence between random variables in the presence of distribution shifts. This versatility underscores the potential impact of our research and its relevance across various domains.


\end{document}